\def\paperID{4638}
\def\confName{ICCV}
\def\confYear{2025}

\def\paperTitle{DiffIR2VR-Zero: Zero-Shot Video Restoration with Diffusion-based Image Restoration Models}

\def\authorBlock{
Chang-Han Yeh$^1$
\quad
Hau-Shiang Shiu$^1$
\quad
Chin-Yang Lin$^1$
\quad
Zhixiang Wang$^2$
\\
Chi-Wei Hsiao$^3$
\quad
Ting-Hsuan Chen$^1$
\quad
Yu-Lun Liu$^1$\vspace{0.75em}
\\
\centerline{$^1$National Yang Ming Chiao Tung University \quad $^2$University of Tokyo \quad $^3$MediaTek Inc.}
}

\newif\ifreview 
\newif\ifarxiv \newcommand{\arxiv}{\arxivtrue}
\newif\ifcamera 
\newif\ifrebuttal 

\arxiv

\pdfoutput=1
\documentclass[10pt,twocolumn,letterpaper]{article}
\ifreview \usepackage[review]{cvpr} \fi
\ifarxiv \usepackage[pagenumbers]{cvpr} \fi
\ifrebuttal \usepackage[rebuttal]{cvpr} \fi
\ifcamera \usepackage{cvpr} \fi

\usepackage{graphicx}	
\usepackage{amsmath}	
\usepackage{amssymb}	
\usepackage{booktabs}
\usepackage{times}
\usepackage{microtype}
\usepackage{epsfig}
\usepackage{caption}
\usepackage{float}
\usepackage{placeins}
\usepackage{color, colortbl}
\usepackage{stfloats}
\usepackage{enumitem}
\usepackage{tabularx}
\usepackage{xstring}
\usepackage{multirow}
\usepackage{xspace}
\usepackage{url}
\usepackage{subcaption}
\usepackage{xcolor}
\usepackage[hang,flushmargin]{footmisc}
\usepackage{makecell}

\ifcamera \usepackage[accsupp]{axessibility} \fi

% ...

\ifarxiv  \fi

\newcommand{\R}[1]{{%
    \textbf{%
        \ifstrequal{#1}{1}{\textcolor{red}{R#1}}{%
        \ifstrequal{#1}{2}{\textcolor{blue}{R#1}}{%
        \ifstrequal{#1}{3}{\textcolor{magenta}{R#1}}{%
        \ifstrequal{#1}{4}{\textcolor{teal}{R#1}}{%
                           \textcolor{cyan}{R#1}%
        }}}}%
    }%
}}

\definecolor{ao(english)}{rgb}{0.0, 0.5, 0.0}

\newcommand\best[1]{\textbf{#1}}

\usepackage{xr-hyper}

\makeatletter
\newcommand*{\addFileDependency}[1]{
  \typeout{(#1)}
  \@addtofilelist{#1}
  \IfFileExists{#1}{}{\typeout{No file #1.}}
}

\makeatother
\newcommand*{\myexternaldocument}[1]{
    \externaldocument{#1}
    \addFileDependency{#1.tex}
    \addFileDependency{#1.aux}
}

\definecolor{cvprblue}{rgb}{0.21,0.49,0.74}
\definecolor{cvprgreen}{rgb}{0,0.8,0.0}
\usepackage[pagebackref,breaklinks,colorlinks,allcolors=cvprblue]{hyperref}
\usepackage[capitalize]{cleveref}
\crefname{section}{Sec.}{Secs.}
\crefname{table}{Table}{Tables}
\crefname{figure}{Fig.}{Figs.}

\ifarxiv \crefname{appendix}{App.}{Apps.}
\else \crefname{appendix}{Suppl.}{Suppls.} \fi

\unless\ifarxiv \myexternaldocument{_supplementary} \fi

\begin{document}
\title{\paperTitle}
\author{\authorBlock}

\twocolumn[{%
\renewcommand\twocolumn[1][]{#1}%
\maketitle
\begin{center}
\centering
\captionsetup{type=figure}
\resizebox{\textwidth}{!} 
{
\includegraphics[width=\textwidth]{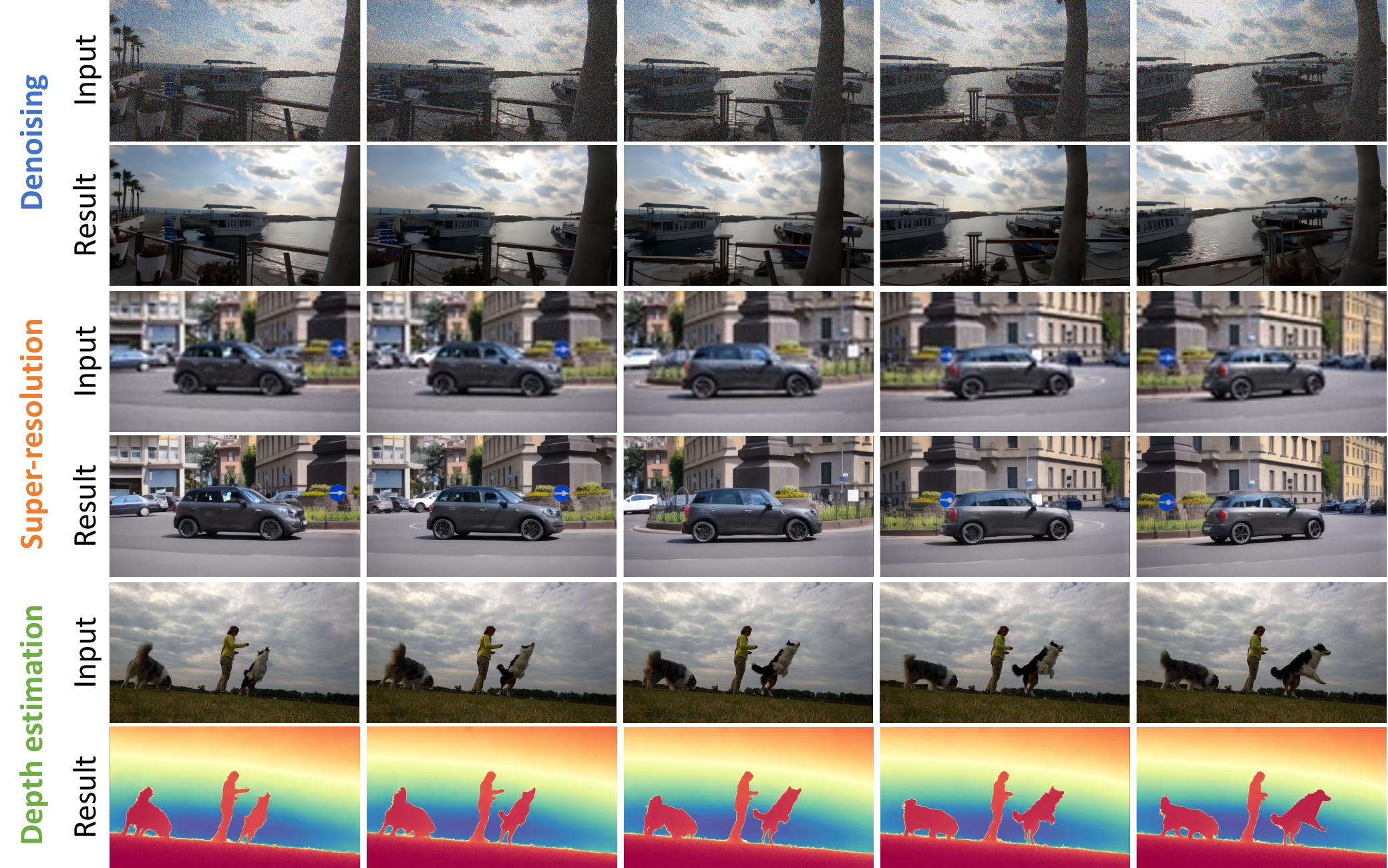}
}
\vspace{-4mm}
\caption{\textbf{Zero-shot temporal-consistent diffusion model for video restoration and beyond.} Given a pre-trained diffusion model for \emph{single-image} restoration, our method generates temporally consistent restored video with fine details \emph{without} {any} further training. Our method applies to other video applications, such as depth estimation. 
}
\label{fig:teaser}
\end{center}
}]

\maketitle

\begin{abstract}

We present DiffIR2VR-Zero, a zero-shot framework that enables any pre-trained image restoration diffusion model to perform high-quality video restoration without additional training. While image diffusion models have shown remarkable restoration capabilities, their direct application to video leads to temporal inconsistencies, and existing video restoration methods require extensive retraining for different degradation types. Our approach addresses these challenges through two key innovations: a hierarchical latent warping strategy that maintains consistency across both keyframes and local frames, and a hybrid token merging mechanism that adaptively combines optical flow and feature matching. Through extensive experiments, we demonstrate that our method not only maintains the high-quality restoration of base diffusion models but also achieves superior temporal consistency across diverse datasets and degradation conditions, including challenging scenarios like 8$\times$ super-resolution and severe noise. Importantly, our framework works with any image restoration diffusion model, providing a versatile solution for video enhancement without task-specific training or modifications.
Please see our project page at \href{https://jimmycv07.github.io/DiffIR2VR_web/}{jimmycv07.github.io/DiffIR2VR\_web}.
\end{abstract}
\section{Introduction}
\label{sec:intro}

\begin{figure}[t]
\centering
\resizebox{1.0\columnwidth}{!} 
{
\includegraphics[width=\textwidth]{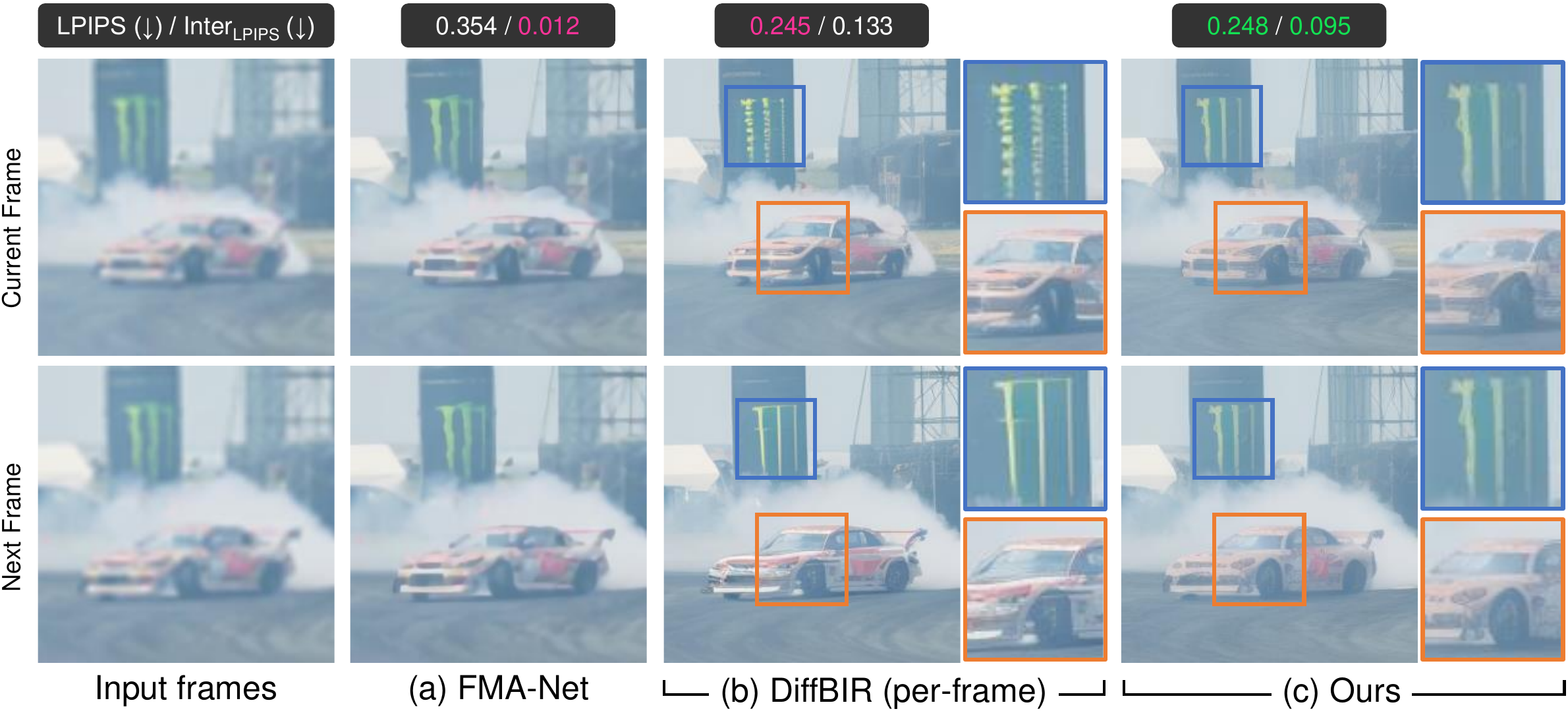}
}
\vspace{-5mm}
\caption{\textbf{4$\times$ video super-resolution results.}
(a) Traditional regression-based methods such as FMA-Net~\citep{youk2024fmanet} are limited to the training data domain and tend to produce blurry results when encountering out-of-domain inputs. (b) Although applying image-based diffusion models such as DiffBIR~\citep{lin2024diffbir} to individual frames can generate realistic details, these details often lack consistency across frames. (c) Our method leverages an image diffusion model to restore videos, achieving both realistic and consistent results \emph{without} any additional training. 
}
\label{fig:motivation}
\end{figure}

Video restoration --- the task of transforming low-quality videos into high-quality ones through denoising, super-resolution, and deblurring --- remains a significant challenge in computer vision. While diffusion models have recently revolutionized image restoration~\citep{xia2023diffir, lin2024diffbir} by generating highly realistic details that surpass traditional regression-based methods (\cref{fig:motivation}(a)), extending these advances to video has proven difficult. Current approaches that directly apply image diffusion models frame-by-frame suffer from severe temporal inconsistencies and flickering artifacts (\cref{fig:motivation}(b)), particularly with Latent Diffusion Models (LDMs).

Existing solutions typically attempt to bridge this gap by fine-tuning image diffusion models with video-specific components like 3D convolution and temporal attention layers. However, these approaches face significant limitations: they require extensive computational resources (\eg, 32 A100-80G GPUs for video upscaling~\citep{zhou2023upscale}), need task-specific retraining, and often struggle to generalize across different degradation types. This creates a pressing need for a more efficient and versatile approach to video restoration.

This paper presents DiffIR2VR-Zero, the first training-free framework that enables zero-shot video restoration using any pre-trained image diffusion model. Unlike previous approaches that require extensive fine-tuning or model modification, our method introduces two key innovations that work in synergy: (i) Hierarchical latent warping that maintains consistency at both global and local scales by intelligently propagating latent features between keyframes and neighboring frames. (ii) Hybrid flow-guided spatial-aware token merging that combines optical flow, similarity matching, and spatial information to achieve robust feature correspondence across frames.
These components work together to enforce temporal consistency in both latent and token spaces while preserving the high-quality restoration capabilities of the underlying image diffusion model (\cref{fig:motivation}(c)). Our approach requires no additional training or fine-tuning, making it immediately applicable to any pre-trained image diffusion model.

As demonstrated in \cref{fig:teaser}, our framework effectively handles a diverse range of restoration tasks, including denoising, super-resolution, and even depth estimation, while maintaining temporal consistency across frames. Extensive experiments show that our method not only achieves state-of-the-art performance in standard scenarios but also excels in extreme cases (such as 8× super-resolution and high-noise denoising) where traditional methods struggle.

While our work builds upon recent advances in video editing with diffusion models~\citep{li2024vidtome,geyer2023tokenflow}, we make several novel contributions that specifically address the challenges of zero-shot video restoration:
\begin{itemize}
\item The first zero-shot framework for adapting any pre-trained image restoration diffusion model to video without additional training, achieving a balance between temporal consistency and detail preservation.
\item A training-free approach that innovatively combines hierarchical latent warping with an improved token merging strategy specifically designed for restoration tasks.
\item State-of-the-art performance across various restoration tasks, demonstrating superior generalization and robustness compared to existing methods, particularly in extreme degradation scenarios.
\end{itemize}

\section{Related Work}
\label{sec:related}

\noindent {\bf Video restoration.}
Video restoration encompasses transforming degraded videos affected by noise, blur, and low resolution into high-quality outputs~\citep{chan2021basicvsr, chan2022basicvsr++, isobe2020video, Li_2023_CVPR, youk2024fmanet}. Unlike single-image restoration~\citep{guo2019toward}, video restoration faces the additional challenge of maintaining temporal consistency across frames. Current approaches primarily rely on motion-based methods using optical flow warping~\citep{huang2022flowformer, pan2020cascaded, shi2023videoflow} or deformable convolutions~\citep{chan2021basicvsr, DeformCNN, wang2019edvr} to align features temporally. Other methods leverage attention mechanisms~\citep{cao2021video, li2020mucan, zamir2022restormer} to model long-range dependencies across frames, while hybrid approaches~\citep{liu2021hybrid, liang2022recurrent} combine multiple techniques to handle complex degradations. However, these methods face several critical limitations: they require extensive paired training data~\citep{chan2022investigating, xie2023mitigating}, assume specific degradation models~\citep{kong2023efficient, li2020mucan}, and need retraining for different degradation levels~\citep{youk2024fmanet}. These constraints significantly limit their real-world applicability and generalization capability.

\noindent {\bf Diffusion models for image restoration.}
Recent advances in diffusion models~\citep{dhariwal2021diffusion, ho2020denoising, rombach2022high} have led to breakthrough improvements in image restoration. Current approaches either train models from scratch~\citep{saharia2022image, xia2023diffir, yue2024resshift}, adapt pre-trained models through guided sampling~\citep{kawar2022denoising}, or fine-tune frozen models with additional layers~\citep{wang2023exploiting, yang2023pixel}, as demonstrated by StableSR and DiffBIR~\citep{lin2024diffbir}. While these methods achieve impressive results for single images, their direct application to video leads to temporal inconsistencies due to the inherent randomness of the diffusion process. Our work uniquely bridges this gap by enabling zero-shot video restoration using these pre-trained image models without any additional training or modification.

\noindent {\bf Video consistency in diffusion models.}
Recent works have explored extending image diffusion models to video tasks~\citep{ho2022video, hu2023lamd, kara2024rave}. Latent-space methods like Rerender-A-Video~\citep{yang2023rerender} use warping and interpolation but struggle with detail preservation in restoration. Token-level approaches like VidToMe~\citep{li2024vidtome} and TokenFlow~\citep{geyer2023tokenflow} often produce over-smoothed results. Our work differs by combining hierarchical latent warping with hybrid correspondence mechanisms, specifically designed for restoration tasks. This enables zero-shot video restoration using any pre-trained image diffusion model, without requiring task-specific training
\begin{figure*}[t]
\centering
\includegraphics[width=\textwidth]{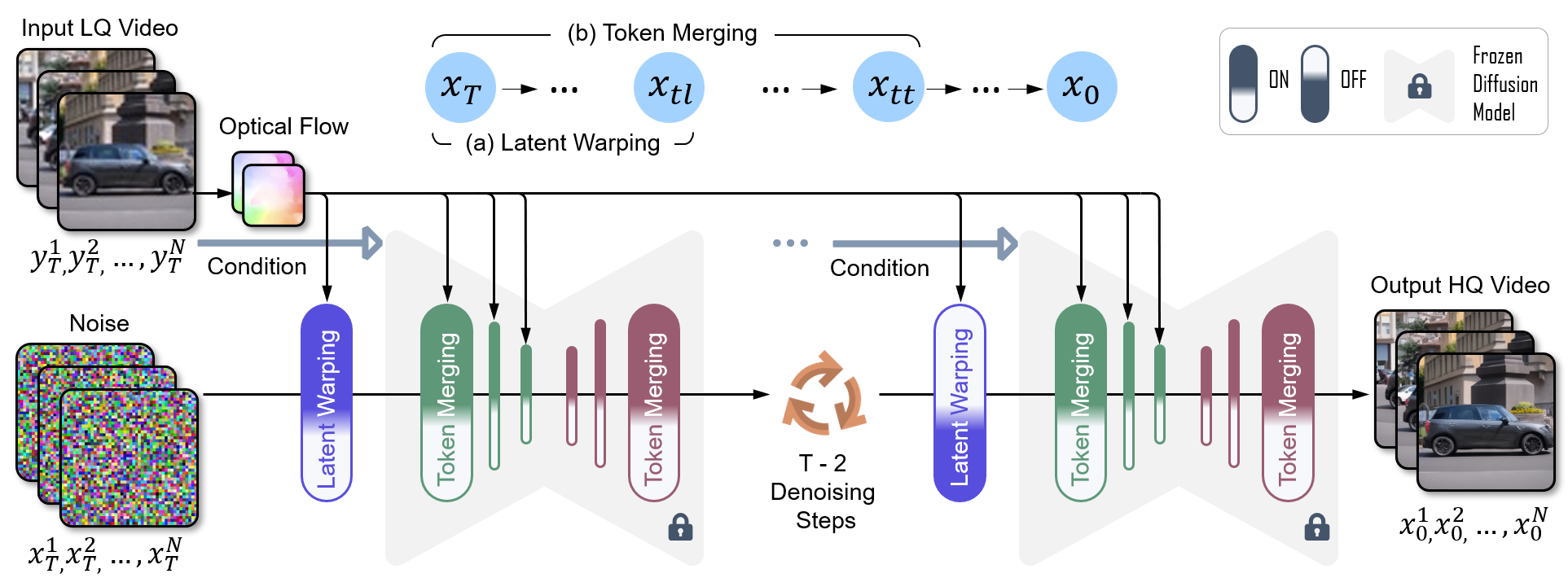}
\vspace{-6mm}
\caption{\textbf{Pipeline of our proposed zero-shot video restoration method.}
We process low-quality (LQ) videos in batches using a diffusion model, with a keyframe randomly sampled within each batch. 
(a) At the beginning of the diffusion denoising process, hierarchical latent warping provides rough shape guidance both globally, through latent warping between keyframes, and locally, by propagating these latents within the batch.
(b) Throughout most of the denoising process, tokens are merged before the self-attention layer. For the downsample blocks, optical flow is used to find the correspondence between tokens, and for the upsample blocks, cosine similarity is utilized. This hybrid flow-guided, spatial-aware token merging accurately identifies correspondences between tokens by leveraging both flow and spatial information, thereby enhancing overall consistency at the token level.
}
\label{fig:pipeline}
\end{figure*}

\section{Method}
\label{sec:method}

Given a low-quality video with $n$ frames ${y^1, y^2, \ldots, y^n}$, our goal is to restore it to high-quality ${x^1, x^2, \ldots, x^n}$ using image-based diffusion models. While direct frame-by-frame application of these models causes temporal inconsistencies due to inherent stochasticity, particularly in extreme degradation cases (\cref{fig:motivation} and \cref{fig:visual_SR}), our method (\cref{fig:pipeline}) addresses this challenge through two key innovations: Hierarchical Latent Warping (\cref{subsec:3_3}) and Hybrid Flow-guided Spatial-aware Token Merging (\cref{subsec:3_4}). In this section, we first introduce the foundational concepts of diffusion models and video token merging, then detail our novel components and their integration.

\subsection{Diffusion Models for Video Editing}
\label{sec:preliminaries}
The forward process of diffusion models progressively adds noise to a clean image $x_0$ over $T$ steps according to:
\begin{equation}
x_t = \sqrt{\alpha_t}x_{t-1} + \sqrt{1-\alpha_t}\epsilon_{t-1} \Rightarrow
x_t = \sqrt{\bar{\alpha}_t} x_0 + \sqrt{1 -\bar{\alpha}_t} \epsilon,
\end{equation}
where $t\sim [1,T]$, $\epsilon_t, \epsilon \sim \mathcal{N}(\mathbf{0}, \mathbf{I})$, and $\bar{\alpha}_t = \prod_{s=1}^{t} \alpha_s$. A UNet-based denoiser $\epsilon\theta$ learns to estimate and remove this noise, with the inverse process gradually denoising $x_t$ to produce $x_0$~\citep{ho2020denoising,song2020denoising}.

Recent video editing techniques like VidToMe~\citep{li2024vidtome} maintain temporal consistency by merging similar tokens within frame chunks. Given a token chunk $\textbf{T} \in \mathbb{R}^{B \times A \times C}$ where $A = w \times h$, tokens are separated into source tokens $\textbf{T}\text{src}$ and a target token $\textbf{T}\text{tar}$. The similarity between tokens is computed as:
\begin{equation}
\begin{aligned}
s(\textbf{T}_\text{src}, \textbf{T}_\text{tar}) = \frac{\textbf{T}_\text{src} \cdot \textbf{T}_\text{tar}}{\left \| \textbf{T}_\text{src} \right \|\left \| \textbf{T}_\text{tar} \right \|}\,,\,\,\,
c  = \max_{\{\textbf{t} \in \textbf{T}_\text{tar}\}}(s(\textbf{T}_\text{src}, \textbf{t})),
\end{aligned}
\end{equation}
where $s(\cdot, \cdot)$ represents cosine similarity and $c$ indicates correspondences. The merging and unmerging operations are defined as:
\begin{equation}
\begin{aligned}
\textbf{T}_\text{merge} = \mathcal{M}(\textbf{T}_\text{src}, \textbf{T}_\text{tar},\;c,\;r)\,,\,\,\,
\textbf{T}_\text{unmerge} = \mathcal{U}(\textbf{T}_\text{merge},\;c)\,.
\end{aligned}
\end{equation}

However, these existing techniques face significant challenges in video restoration. Early-stage denoising produces noisy latents that make traditional similarity measures unreliable, especially in UNet's downsample blocks (\cref{fig:correspondence}). Additionally, focusing solely on frame-to-frame consistency misses global-local coherence, while aggressive token merging can lead to over-smoothing.

\begin{figure}[t]
\centering
\includegraphics[width=0.95\columnwidth]{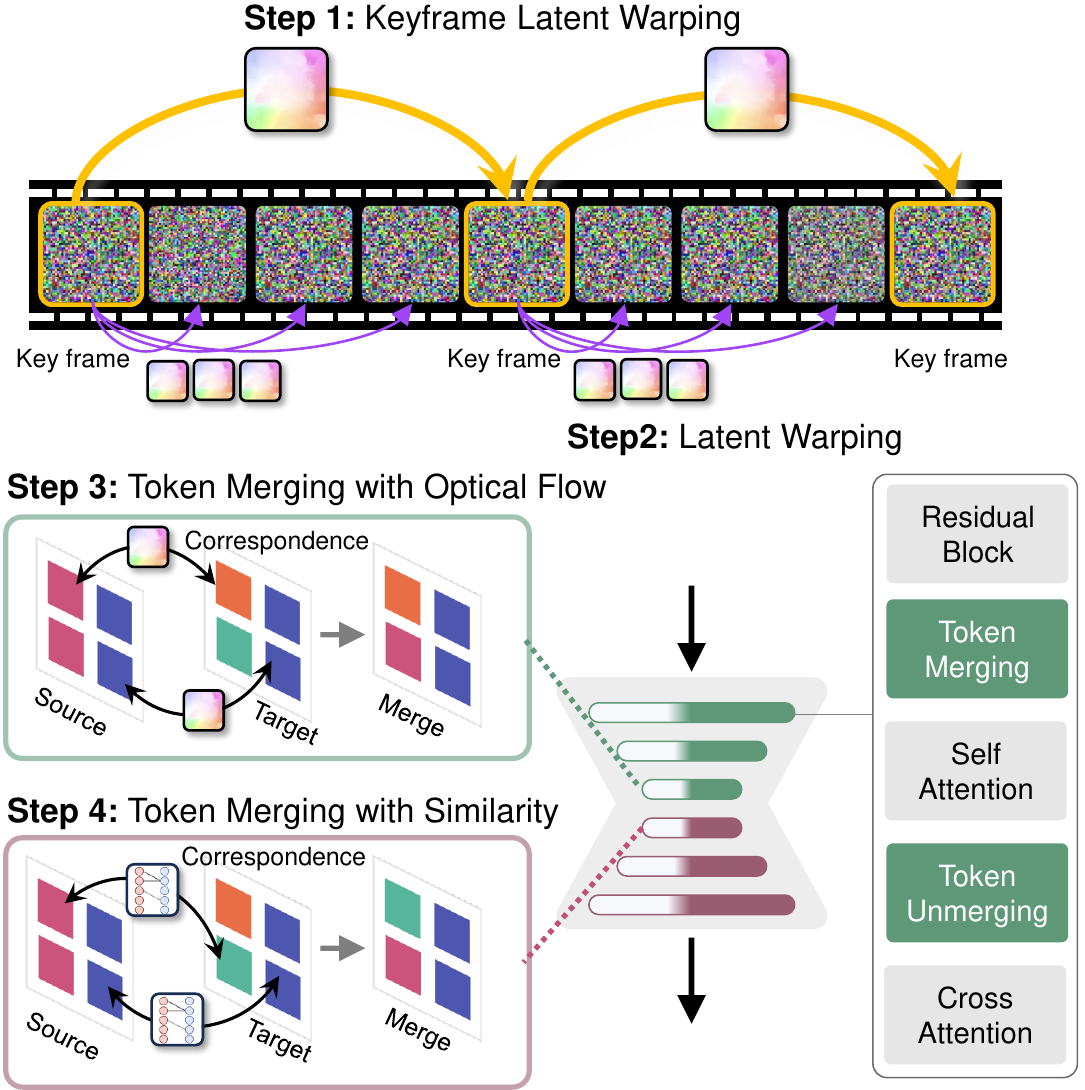}
\caption{\textbf{An illustration of our key modules.}
Without requiring any training, these modules can achieve coherence across frames by enforcing temporal stability in both latent and token space. Hierarchical latent warping provides global and local shape guidance; Hybrid spatial-aware token merging before the self-attention layer improves temporal consistency by matching similar tokens using optical flow in the down blocks and cosine similarity in the up blocks of the UNet.
}
\label{fig:component}
\end{figure}

\subsection{Hierarchical Latent Warping} \label{sec:hierarchical}
\label{subsec:3_3}
Our key innovation in maintaining temporal consistency begins with a hierarchical latent warping module that operates at two distinct levels in the latent space. The first level handles global consistency across the video by warping between keyframes, while the second level ensures local consistency by propagating these warped latents within each processing batch. This hierarchical approach provides essential shape guidance at both global and local scales, as illustrated in \cref{fig:component} (upper part).
For a given latent $\hat{x}^{i}{t \rightarrow 0}$ predicted for the $i^{th}$ keyframe at denoising step $t$, we first establish global consistency through keyframe warping:
\begin{equation}
\hat{x}^{i}_{t \rightarrow 0} \leftarrow M_{ji} \cdot \hat{x}^{i}_{t \rightarrow 0} + \left(1 - M_{ji}\right) \cdot \mathcal{W}(\hat{x}^{j}_{t \rightarrow 0}, f_{ji}),
\end{equation}
where $j = i - 1$ represents the previous keyframe. The optical flow field $f_{ji}$ and occlusion mask $M_{ji}$ are computed from low-quality frames $lq_j$ to $lq_i$ using GMFlow~\citep{xu2022gmflow}. This formulation allows us to blend the original latent features with warped features from the previous keyframe, weighted by the occlusion mask to handle regions where warping may be unreliable.

Following global alignment, we propagate these warped latents to all frames within the current batch, establishing local consistency. Unlike previous approaches that rely solely on frame-to-frame warping, our hierarchical strategy ensures that corresponding points maintain similar latent representations across both temporal scales from the early stages of the denoising process. This comprehensive approach to temporal consistency proves particularly effective in handling complex motion patterns and maintaining coherent structure across the entire video sequence.
To address potential warping errors in severely degraded regions, we incorporate forward-backward consistency checks and selective feature propagation. This makes our approach more robust to optical flow failures and occlusions compared to single-scale warping methods. Our experiments demonstrate that this hierarchical strategy significantly reduces temporal artifacts while preserving fine details in the restored video.

\subsection{Hybrid Flow-guided Spatial-aware Token Merging} \label{sec:hybrid}
\label{subsec:3_4}
While latent manipulation effectively maintains consistency in early stages, it can produce blurry results when applied during later denoising stages. To address this limitation, we introduce a hybrid flow-guided spatial-aware token merging approach that operates in the semantically rich token space, achieving both temporal consistency and detail preservation.

\begin{figure}[t]
\centering
\includegraphics[width=\columnwidth]{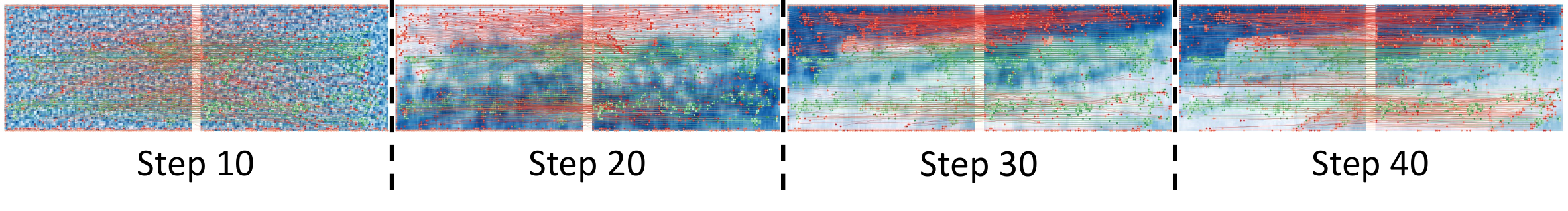}
\vspace{-7mm}
\caption{\textbf{Token correspondences (\textcolor{red}{cosine similarity} and \textcolor{ao(english)}{optical flow}) across denoising steps.}
Early on (\emph{e.g.}, step 10), optical flow guides better due to noisy latents. Later (\emph{e.g.}, step 40), similarity and flow focus on different regions, showcasing the benefit of our hybrid approach for effective token merging throughout denoising.
}
\label{fig:correspondence}
\end{figure}

\noindent {\bf Flow guidance and spatial-awareness.}
Our key insight is that different correspondence mechanisms are optimal at different stages of the denoising process. In early stages, when latent representations are noisy, traditional cosine similarity measures become unreliable, particularly in the UNet's downsample blocks (\cref{fig:correspondence}, top). During these stages, optical flow computed from low-resolution inputs provides more reliable guidance. As denoising progresses (\emph{e.g.}, steps 30-40), flow-based and similarity-based methods often identify complementary correspondences (\cref{fig:correspondence}, bottom), motivating our hybrid approach.
For downsample blocks, we employ flow-guided correspondence with a forward-backward consistency check to ensure reliability:
\begin{align} \label{eq:flow_weight}
\sigma = e^{(-\left\| f_{\text{src} \rightarrow \text{tar}}(X(\textbf{T}_\text{src})) + f_{\text{tar} \rightarrow \text{src}}\left(X(\textbf{T}_\text{src}) + f_{\text{src} \rightarrow \text{tar}}(X(\textbf{T}_\text{src}))\right) \right\|_2^2)},
\end{align}
where $\sigma$ represents the confidence score, $X(\textbf{T}_\text{src})$ denotes the spatial location of source token $\textbf{T}_\text{src}$, and $f_{\text{src} \rightarrow \text{tar}}$, $f_{\text{tar} \rightarrow \text{src}}$ represent forward and backward optical flows. This confidence score guides the token merging process:
\begin{equation}
\textbf{T}_\text{merge} = \mathcal{M}(\textbf{T}_\text{src},\;\textbf{T}_\text{tar}, \;f_{\text{src} \rightarrow \text{tar}},\; \sigma,\;r).
\end{equation}
For upsample blocks, we enhance cosine similarity matching with spatial awareness to prevent mismatches in uniform texture regions (\emph{e.g.}, sky, grass). We weight similarity scores based on spatial proximity:
\begin{equation}
s_{ij}^{\prime} = s_{ij} \cdot e^{-\tau}\,, \, \, \text{with}\,\, \tau = \left\lfloor \left[{\|X(i) - X(j)\|_2^2}\right]/{R} \right\rfloor\,,
\end{equation}
where $X(i)$, $X(j)$ are token spatial locations and $R$ defines the radius of influence. To prevent padding artifacts, we remove padding before merging and restore it after unmerging.

\noindent {\bf Merging ratio annealing.}
To maintain high-quality results throughout the denoising process, we implement ratio annealing that gradually reduces the merging ratio:
\begin{equation}
r_{i} = r \cdot \cos\left(\frac{\pi}{2} \cdot \text{max} \left( \text{min}\left(\delta \cdot  \frac{i - i_{\text{beg}}}{i_{\text{end}} - i_{\text{beg}}}, 1 \right), 0\right)\right),
\end{equation}
where $i_{\text{beg}}$ and $i_{\text{end}}$ define the annealing period, and $\delta$ controls the annealing speed. This annealing strategy helps balance temporal consistency with detail preservation, avoiding the over-smoothing common in regression-based methods while maintaining better temporal coherence than per-frame processing (\cref{fig:motivation}).

\subsection{Scheduling}
\label{sec:schedule}
Our method orchestrates different components through the denoising process. In early stages (steps 1-10), hierarchical latent warping establishes global and local consistency by warping between keyframes and propagating features within batches. During the main denoising phase (steps 10-40), we switch to hybrid spatial-aware token merging before each attention layer. This mechanism adapts based on network depth: optical flow guides downsample blocks while similarity matching handles upsample blocks. Throughout the process, our annealing schedule gradually reduces the merging ratio, starting aggressively for consistency and becoming more conservative to preserve details.

\section{Experiments}
We conduct extensive experiments to evaluate our zero-shot video restoration framework across different tasks, datasets, and degradation levels. We focus particularly on challenging scenarios that test both restoration quality and temporal consistency.

\noindent {\bf Datasets and evaluation protocol.}
For video super-resolution, we evaluate on three standard benchmarks: REDS4~\citep{nah2019ntire}, Vid4~\citep{liu2013bayesian}, and DAVIS~\citep{7780454}. We test at multiple upscaling factors ($\times$4 and $\times$8) using the realistic degradation model from RealBasicVSR~\citep{chan2022investigating} to simulate real-world conditions. For video denoising, we use REDS30~\citep{nah2019ntire} and Set8~\citep{9156652}, testing across a range of noise levels (std. = 50, 75, 100, 150) as well as random noise in the [50, 100] range to evaluate robustness to varying degradation severity.

\noindent {\bf Evaluation metrics.}
Our evaluation considers both perceptual quality and temporal consistency through complementary metrics: (1) For perceptual quality, we use LPIPS for assessing visual realism, alongside traditional PSNR and SSIM metrics. (2) For temporal consistency, we employ warping error ($E_{\text{warp}}$), frame interpolation error, and our proposed interpolation LPIPS. The latter metric extends the interpolation error concept from~\citep{li2024vidtome} by using LPIPS to better capture perceptual temporal consistency, measuring how well interpolated frames match the actual frames.

\noindent {\bf Implementation details.}
We implement our framework using PyTorch and conduct experiments on an NVIDIA RTX 4090 GPU. To demonstrate the versatility of our approach, we apply it to two different image restoration diffusion models: DiffBIR~\citep{lin2024diffbir} and the SD$\times$4 upscaler~\citep{sdx4}. For achieving 8$\times$ super-resolution with models limited to 4$\times$ upscaling, we cascade the process twice followed by bicubic downsampling. While this approach works well with DiffBIR, we note that memory constraints prevent its application with SDx4 on the REDS dataset due to the larger image sizes involved.

\subsection{Comparisons with State-of-the-Art Methods}

We conduct comprehensive comparisons with leading methods across different video restoration tasks, examining both traditional learning-based approaches and recent diffusion-based methods.

\begin{table*}[t]
\caption{\textbf{Quantitative comparisons of video super-resolution on the DAVIS~\citep{perazzi2016benchmark}, Vid4~\citep{liu2013bayesian} and REDS4~\citep{nah2019ntire} datasets.} The best and second performances are marked in \textcolor{red}{red} and \textcolor{blue}{blue}, respectively. $E_{\text{warp}}^*$ denotes $E_{\text{warp}} (\times 10^{-3})$ and  $E_{\text{inter}}$,
LPIPS$_{\text{inter}}$ denotes interpolation error and LPIPS. - indicates out-of-memory.}
\label{tab:SR}
\vspace{-3mm}
\centering
\resizebox{\textwidth}{!}
{
\begin{tabular}{l|r|rrr|rrrrrr}
\toprule 
& & BasicVSR++~\cite{chan2022basicvsr++} & RVRT~\cite{liang2022recurrent} & FMA-Net~\cite{youk2024fmanet} & VidToMe~\cite{li2024vidtome} & \multicolumn{2}{c}{SD $\times$4~\cite{sdx4}} & \multicolumn{2}{c}{DiffBIR~\cite{lin2024diffbir} (ECCV 2024)}\\\cmidrule(lr){7-8}\cmidrule(lr){9-10}
& Metrics & (CVPR 2022) & (NeurIPS 2022) & (CVPR 2024) & (CVPR 2024) & Frame & {{Ours {\small{(Improve)}}}} & {Frame} & {{Ours {\small{(Improve)}}}} \\
\midrule
\multirow{6}{*}{\rotatebox{90}{DAVIS $\times$4}} & PSNR $\uparrow$ & \textcolor{blue}{26.576} & \textcolor{red}{26.595} & {25.215} & 23.014 & 23.504 & 23.843 {\small\textbf{(+0.339)}} & 23.780 & {24.182} {\small\textbf{(+0.402)}} \\
 & SSIM $\uparrow$ & \textcolor{blue}{0.743} & \textcolor{red}{0.744} & {0.727} & 0.566  & 0.584  & 0.618 {\small\textbf{(+0.034)}} & 0.601  & {0.621} {\small\textbf{(+0.020)}} \\
 & LPIPS $\downarrow$ & 0.383 & 0.388 & 0.347  & 0.405 & 0.277  & 0.272 {\small\textbf{(-0.005)}} & \textcolor{blue}{0.264} & \textcolor{red}{0.262} {\small\textbf{(-0.002)}} \\
 & $E_{\text{warp}}^* \downarrow$ & \textcolor{red}{0.090} & \textcolor{red}{0.090} & \textcolor{blue}{0.186} & 0.520  & 0.912  & 0.745 {\small\textbf{(-0.167)}} & {0.654} & {0.474} {\small\textbf{(-0.180)}} \\
 & $E_{\text{inter}} \downarrow$ & \textcolor{red}{9.115} & \textcolor{blue}{9.135} & {11.558} & {13.676} & 18.125 & 17.431 {\small\textbf{(-0.694)}} & 16.529 & 14.666 {\small\textbf{(-1.863)}} \\
 & LPIPS$_{\text{inter}}\downarrow$ & \textcolor{red}{0.058} & \textcolor{red}{0.058} & \textcolor{blue}{0.078} & 0.329 &  0.292  & 0.274 {\small\textbf{(-0.018)}} & 0.266  & 0.232 {\small\textbf{(-0.034)}} \\
\midrule
\multirow{6}{*}{\rotatebox{90}{DAVIS $\times$8}} & PSNR $\uparrow$ & \textcolor{blue}{24.301} & \textcolor{red}{24.504} & {22.690} & 22.097 & 20.268 & 20.519 {\small\textbf{(+0.251)}} & 21.964 & {22.331} {\small\textbf{(+0.367)}} \\
 & SSIM $\uparrow$ & \textcolor{blue}{0.631} & \textcolor{red}{0.638} & {0.594} & 0.513 & 0.446  & 0.424 {\small(-0.022)} & 0.502  & {0.519} {\small\textbf{(+0.017)}} \\
 & LPIPS $\downarrow$ & 0.518 & 0.560 & 0.528  & 0.554  & 0.470  & 0.434 {\small\textbf{(-0.036)}} & \textcolor{red}{0.362} & \textcolor{blue}{0.367} {\small(+0.005)} \\
 & $E_{\text{warp}}^* \downarrow$ & \textcolor{blue}{0.132} & \textcolor{red}{0.127} & {0.351} & {0.440}  & 2.199  & 1.759 {\small\textbf{(-0.440)}} & 0.964 & 0.699 {\small\textbf{(-0.265)}} \\
 & $E_{\text{inter}} \downarrow$ & \textcolor{blue}{9.882} & \textcolor{red}{9.725} & {13.978} & {12.624} &  24.496 & 21.746 {\small\textbf{(-2.750)}} & 17.981 & 15.853 {\small\textbf{(-2.128)}} \\
 & LPIPS$_{\text{inter}}\downarrow$ & \textcolor{blue}{0.088} & \textcolor{red}{0.081} & {0.132} & 0.388  & 0.457  & 0.442 {\small\textbf{(-0.015)}} & 0.372  & {0.333} {\small\textbf{(-0.039)}} \\
\midrule
\multirow{6}{*}{\rotatebox{90}{REDS4 $\times$4}} & PSNR $\uparrow$ & \textcolor{blue}{27.227} & \textcolor{red}{27.244} & {25.829} & 23.134 & 24.189 & 24.226 {\small\textbf{(+0.037)}} & 24.679 & {25.118} {\small\textbf{(+0.439)}} \\
 & SSIM $\uparrow$ & \textcolor{red}{0.781} & \textcolor{red}{0.781} & \textcolor{blue}{0.761} & 0.589 & 0.638 & 0.641 {\small\textbf{(+0.003)}} & 0.657 & {0.683} {\small\textbf{(+0.026)}} \\
 & LPIPS $\downarrow$ & 0.369 & 0.374 & 0.327 & 0.357 & 0.247 & 0.242 {\small\textbf{(-0.005)}} & \textcolor{red}{0.211} & \textcolor{blue}{0.222} {\small{(+0.011)}} \\
 & $E_{\text{warp}}^* \downarrow$ & \textcolor{blue}{0.134} & \textcolor{red}{0.133} & {0.392} & 0.579 & 0.817 & 0.811 {\small\textbf{(-0.006)}} & 0.704 & {0.499} {\small\textbf{(-0.205)}} \\
 & $E_{\text{inter}} \downarrow$ & \textcolor{red}{15.799} & \textcolor{blue}{15.838} & {19.014} & {17.869} & 22.906 & 22.889 {\small\textbf{(-0.017)}} & 22.305 & 20.130 {\small\textbf{(-2.175)}} \\
 & LPIPS$_{\text{inter}}\downarrow$ & \textcolor{blue}{0.106} & \textcolor{red}{0.101} & {0.133} & 0.356 & 0.295 & 0.281 {\small\textbf{(-0.014)}} & 0.271 & {0.221} {\small\textbf{(-0.050)}} \\
\midrule
\multirow{6}{*}{\rotatebox{90}{REDS4 $\times$8}} & PSNR $\uparrow$ & \textcolor{blue}{26.109} & \textcolor{red}{26.226} & {22.842} & 21.894 & 20.601 & 20.622 {\small\textbf{(+0.021)}} & 22.479 & {22.961} {\small\textbf{(+0.482)}} \\
 & SSIM $\uparrow$ & \textcolor{blue}{0.719} & \textcolor{red}{0.726} & {0.644} & 0.532 & 0.519 & 0.506 {\small{(-0.013)}} & 0.559 & {0.590} {\small\textbf{(+0.031)}} \\
 & LPIPS $\downarrow$ & 0.436 & 0.431 & 0.423 & 0.538 & 0.386 & 0.367 {\small\textbf{(-0.019)}} & \textcolor{blue}{0.311} & \textcolor{red}{0.306} {\small\textbf{(-0.005)}} \\
 & $E_{\text{warp}}^* \downarrow$ & \textcolor{red}{0.127} & \textcolor{blue}{0.129} & 0.753 & {0.423} & 1.928 & 1.735 {\small\textbf{(-0.247)}} & 0.828 & {0.551} {\small\textbf{(-0.277)}} \\
 & $E_{\text{inter}} \downarrow$ & \textcolor{blue}{15.753} & 15.822 & 21.519 & \textcolor{red}{15.502} & 26.886 & 25.503 {\small\textbf{(-1.383)}} & 21.76 & {19.382} {\small\textbf{(-2.378)}} \\
 & LPIPS$_{\text{inter}}\downarrow$ & \textcolor{red}{0.099} & \textcolor{red}{0.099} & \textcolor{blue}{0.159} & 0.412 & 0.370 & 0.388 {\small{(+0.018)}} & 0.351 & {0.287} {\small\textbf{(-0.064)}} \\
\midrule
\multirow{6}{*}{\rotatebox{90}{REDS4 $\times$16}} & PSNR $\uparrow$ & \textcolor{blue}{23.579} & \textcolor{red}{23.715} & {21.569} & 20.520 & 18.706 & 18.858 {\small\textbf{(+0.152)}} & 20.124 & {20.712} {\small\textbf{(+0.588)}} \\
 & SSIM $\uparrow$ & \textcolor{blue}{0.616} & \textcolor{red}{0.621} & {0.570} & 0.483 & 0.461 & 0.410 {\small{(-0.051)}} & 0.461 & {0.509} {\small\textbf{(+0.048)}} \\
 & LPIPS $\downarrow$ & 0.600 & 0.596 & 0.565 & 0.697 & 0.612 & 0.562 {\small\textbf{(-0.050)}} & \textcolor{blue}{0.446} & \textcolor{red}{0.438} {\small\textbf{(-0.008)}} \\
 & $E_{\text{warp}}^* \downarrow$ & \textcolor{red}{0.084} & \textcolor{blue}{0.085} & {0.619} & {0.296} & 2.664 & 2.030 {\small\textbf{(-0.634)}} & 1.168 & 0.665 {\small\textbf{(-0.503)}} \\
 & $E_{\text{inter}} \downarrow$ & \textcolor{blue}{14.069} & 14.267 & 18.758 & \textcolor{red}{12.945} & 28.478 & 24.000 {\small\textbf{(-4.478)}} & 21.33 & {17.731} {\small\textbf{(-3.599)}} \\
 & LPIPS$_{\text{inter}}\downarrow$ & \textcolor{red}{0.088} & \textcolor{red}{0.088} & \textcolor{blue}{0.139} & 0.417 & 0.559 & 0.493 {\small\textbf{(-0.066)}} & 0.444 & {0.358} {\small\textbf{(-0.086)}} \\
\midrule
\multirow{6}{*}{\rotatebox{90}{Vid4 $\times$4}} & PSNR $\uparrow$ & 23.142 & \textcolor{blue}{23.160} & \textcolor{red}{23.209} & 19.622 & 20.047 & 20.134 {\small\textbf{(+0.087)}} & 20.687 & {21.226} {\small\textbf{(+0.539)}} \\
 & SSIM $\uparrow$ & 0.667 & \textcolor{blue}{0.669} & \textcolor{red}{0.679} & 0.425 & 0.478 & 0.473 {\small{(-0.005)}} & 0.497 & {0.525} {\small\textbf{(+0.028)}} \\
 & LPIPS $\downarrow$ & 0.418 & 0.423 & 0.375 & 0.491 & 0.343 & 0.331 {\small\textbf{(-0.012)}} & \textcolor{blue}{0.329} & \textcolor{red}{0.326} {\small\textbf{(-0.003)}} \\
 & $E_{\text{warp}}^* \downarrow$ & \textcolor{blue}{0.173} & \textcolor{red}{0.167} & {0.203} & 0.687 & 1.502 & 1.397 {\small\textbf{(-0.105)}} & 1.156 & {0.677} {\small\textbf{(-0.479)}} \\
 & $E_{\text{inter}} \downarrow$ & \textcolor{red}{3.398} & \textcolor{blue}{3.399} & {4.442} & 11.754 & 17.234 & 16.921 {\small\textbf{(-0.313)}} & 15.478 & {11.316} {\small\textbf{(-4.162)}} \\
 & LPIPS$_{\text{inter}}\downarrow$ & \textcolor{red}{0.015} & \textcolor{red}{0.015} & \textcolor{blue}{0.026} & 0.337 & 0.275 & 0.271 {\small\textbf{(-0.004)}} & 0.265 & {0.198} {\small\textbf{(-0.067)}} \\
\midrule
\multirow{6}{*}{\rotatebox{90}{Vid4 $\times$8}} & PSNR $\uparrow$ & \textcolor{blue}{21.601} & \textcolor{red}{21.707} & {21.033} & 18.811 & 17.813 & 17.992 {\small\textbf{(+0.179)}} & 18.636 & {19.304} {\small\textbf{(+0.668)}} \\
 & SSIM $\uparrow$ & \textcolor{blue}{0.546} & \textcolor{red}{0.552} & {0.521} & 0.372 & 0.345 & 0.307 {\small{(-0.038)}} & 0.367 & {0.406} {\small\textbf{(+0.039)}} \\
 & LPIPS $\downarrow$ & 0.535 & 0.528 & 0.514 & 0.654 & 0.507 & 0.484 {\small\textbf{(-0.023)}} & \textcolor{blue}{0.440} & \textcolor{red}{0.435} {\small\textbf{(-0.005)}} \\
 & $E_{\text{warp}}^* \downarrow$ & \textcolor{red}{0.139} & \textcolor{blue}{0.151} & {0.221} & {0.477} & 2.523 & 1.972 {\small\textbf{(-0.551)}} & 1.524 & 0.767 {\small\textbf{(-0.757)}} \\
 & $E_{\text{inter}} \downarrow$ & \textcolor{red}{3.170} & \textcolor{blue}{3.193} & {5.269} & {9.942} & 22.881 & 19.970 {\small\textbf{(-2.911)}} & 18.112 & 12.281 {\small\textbf{(-5.831)}} \\
 & LPIPS$_{\text{inter}}\downarrow$ & \textcolor{red}{0.011} & \textcolor{red}{0.011} & \textcolor{blue}{0.032} & 0.393 & 0.423 & 0.419 {\small\textbf{(-0.004)}} & 0.395 & {0.294} {\small\textbf{(-0.101)}} \\
\bottomrule
\end{tabular}
}
\end{table*}
\begin{figure}[t]
\centering
{
\includegraphics[width=\columnwidth]{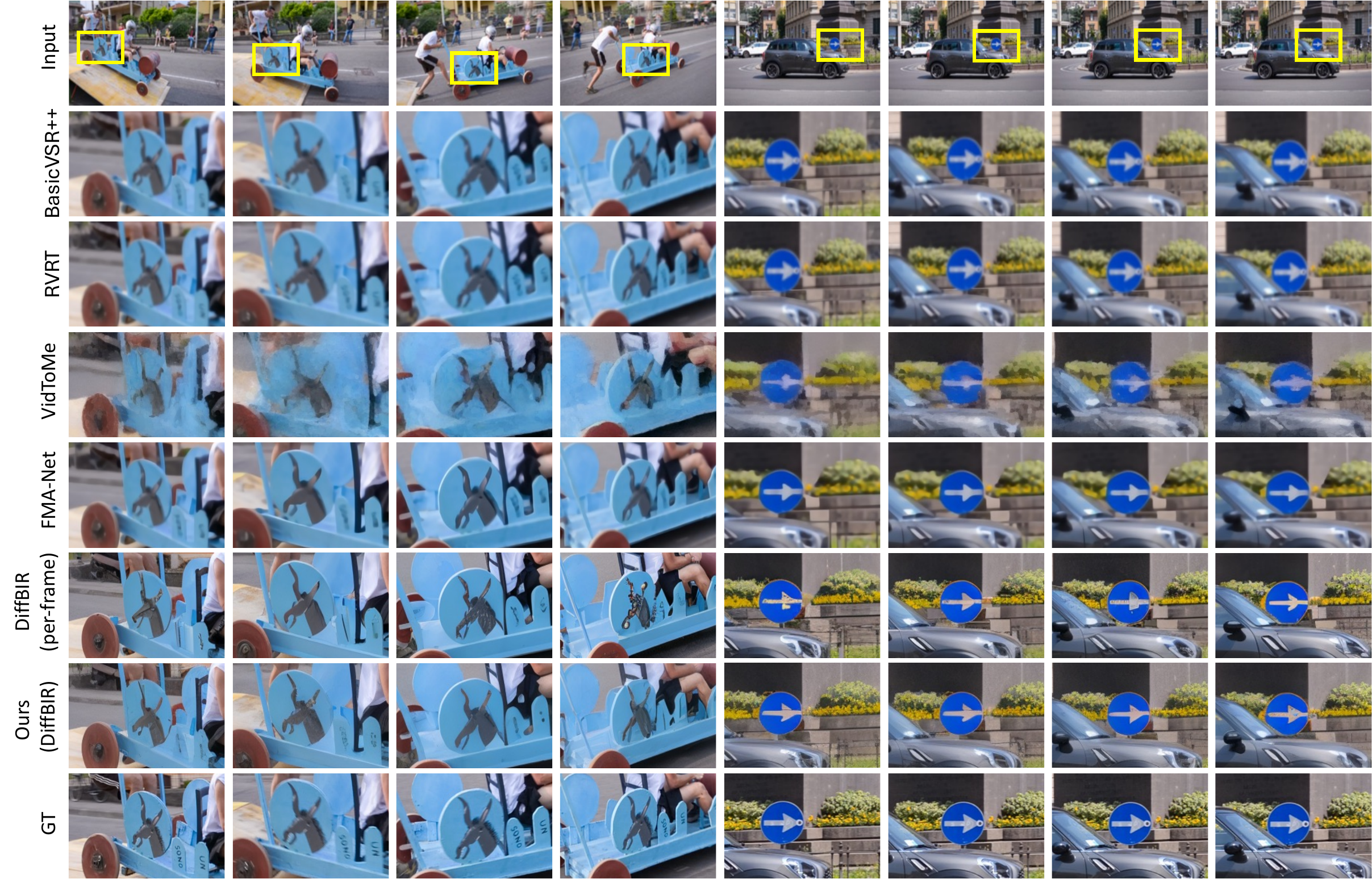}
}
\vspace{-7mm}
\caption{\textbf{Qualitative comparisons on 4$\times$ video super-resolution.} 
As shown in the first row, the low-quality input lacks almost all details. In the zoomed-in patches, our method produces clearer and more consistent results.
}
\label{fig:visual_SR}
\end{figure}
\begin{figure}[t]
\centering
{
\includegraphics[width=\columnwidth]{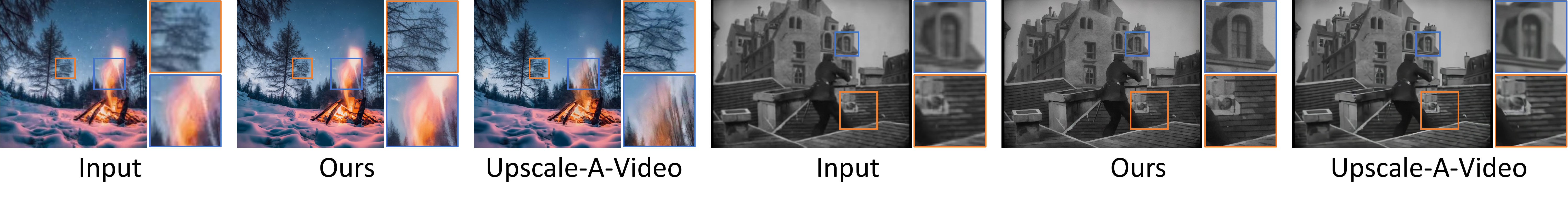}
}
\vspace{-8mm}
\caption{\textbf{Qualitative comparisons with Upscale-A-Video~\citep{zhou2023upscale} on 4$\times$ video SR.}
}
\label{fig:visual_SR_upscale_a_video}
\end{figure}

\noindent {\bf Video super-resolution.}
We compare against state-of-the-art methods including BasicVSR++~\citep{chan2022basicvsr++}, RVRT~\citep{liang2022recurrent}, and FMA-Net~\citep{youk2024fmanet}. Additionally, we evaluate against VidToMe~\citep{li2024vidtome} applied to our base models and attempted comparisons with Upscale-A-Video~\citep{zhou2023upscale}, though hardware limitations (48GB A6000 GPU) prevented direct comparison due to memory constraints.

Quantitative results in \cref{tab:SR} reveal several key findings: regression-based methods like FMA-Net struggle with severe degradation and large motion, while VidToMe achieves temporal consistency but produces overly smooth results with poor visual quality. Our method uniquely maintains both the high-quality generation capabilities of the base diffusion model and strong temporal consistency.

Visual comparisons in \cref{fig:visual_SR} further demonstrate these differences. FMA-Net's results show clear limitations when dealing with domain gaps between training and testing conditions. While per-frame application of DiffBIR~\citep{lin2024diffbir} and SD×4 upscaler~\citep{sdx4} produces sharp details, the results suffer from temporal inconsistencies and jittering. Our approach successfully combines high-fidelity restoration with temporal stability. For comparisons with Upscale-A-Video on their test cases (\cref{fig:visual_SR_upscale_a_video}), our method achieves superior detail preservation, leveraging pre-trained diffusion priors more effectively than their fine-tuning approach.

\begin{table}[t]
\caption{\textbf{Quantitative comparisons of video denoising of various noise levels on the REDS30 and Set8~\citep{Set8} dataset.} The best and second performances are marked in \textcolor{red}{red} and \textcolor{blue}{blue}, respectively. $E_{\text{warp}}^*$ denotes $E_{\text{warp}} (\times 10^{-3})$ and  $E_{\text{inter}}$,
LPIPS$_{\text{inter}}$ denotes interpolation error and LPIPS.}
\vspace{-3mm}
\label{tab:denoise}
\centering
\resizebox{\columnwidth}{!}
{
\begin{tabular}{c|r|rrrr}
\toprule
& & {VidToMe~\cite{li2024vidtome}} & {Shift-Net~\cite{yan2018shift}} & \multicolumn{2}{c}{DiffBIR~\cite{lin2024diffbir} (ECCV 2024)}\\\cmidrule(lr){5-6}
 $\sigma$ & Metrics  & (CVPR 2024) & (CVPR 2023) & {Frame} & {Ours {\small{(Improve)}}} \\
\midrule
 \multirow{6}{*}{\rotatebox{90}{REDS30 75}} & PSNR $\uparrow$ & 22.671 & 21.033 & \textcolor{red}{24.585} & \textcolor{blue}{24.520} {\small{(-0.065)}} \\
 & SSIM $\uparrow$ & 0.559  & 0.381  & \textcolor{red}{0.649} & \textcolor{red}{0.649} {\small{(+0.000)}}  \\
 & LPIPS $\downarrow$ & 0.397  & 0.735  & \textcolor{blue}{0.276} & \textcolor{red}{0.275} {\small\textbf{(-0.001)}}  \\
 & $E_{\text{warp}}^* \downarrow$ & \textcolor{blue}{0.727} & 0.765  & 0.751 & \textcolor{red}{0.706} {\small\textbf{(-0.045)}}  \\
 & $E_{\text{inter}} \downarrow$ & \textcolor{red}{18.440} & 21.751 & 21.798 & \textcolor{blue}{21.166} {\small\textbf{(-0.632)}} \\
 & LPIPS$_{\text{inter}}\downarrow$ & 0.375  & 0.501  & \textcolor{red}{0.275} & \textcolor{red}{0.264} {\small\textbf{(-0.011)}}  \\\midrule
 \multirow{6}{*}{\rotatebox{90}{REDS30 100}} & PSNR $\uparrow$ & 22.588 & 22.573 & \textcolor{blue}{24.524} & \textcolor{red}{24.534} {\small\textbf{(+0.010)}} \\
 & SSIM $\uparrow$ & 0.557  & 0.484   & \textcolor{blue}{0.648} & \textcolor{red}{0.652} {\small\textbf{(+0.004)}}  \\
 & LPIPS $\downarrow$ & 0.404  & 0.518 & \textcolor{blue}{0.275} & \textcolor{red}{0.271} {\small\textbf{(-0.004)}}  \\
& $E_{\text{warp}}^* \downarrow$ & \textcolor{blue}{0.733}  & 1.126 & 0.763 & \textcolor{red}{0.696} {\small\textbf{(-0.067)}}  \\
& $E_{\text{inter}} \downarrow$ & \textcolor{red}{18.370} & 23.424 & 21.835 & \textcolor{red}{20.639} {\small\textbf{(-1.196)}} \\
& LPIPS$_{\text{inter}}\downarrow$ & 0.380  & 0.375 & \textcolor{blue}{0.281} & \textcolor{red}{0.267} {\small\textbf{(-0.014)}}  \\ \midrule
\multirow{6}{*}{\rotatebox{90}{REDS30 random}} & PSNR $\uparrow$ & 22.348 & 21.113 & \textcolor{red}{24.579} & \textcolor{blue}{24.508} {\small{(-0.071)}} \\
& SSIM $\uparrow$ & 0.546  & 0.386 & \textcolor{red}{0.650} & \textcolor{blue}{0.649} {\small{(-0.001)}}  \\
& LPIPS $\downarrow$ & 0.429 & 0.728 & \textcolor{blue}{0.276} & \textcolor{red}{0.270} {\small\textbf{(-0.006)}}  \\
& $E_{\text{warp}}^* \downarrow$ & \textcolor{red}{0.681} & 1.896 & 0.755 & \textcolor{blue}{0.713} {\small\textbf{(-0.042)}}  \\
& $E_{\text{inter}} \downarrow$ & \textcolor{red}{17.608} & 27.565 & 21.743 & \textcolor{blue}{21.140} {\small\textbf{(-0.603)}} \\
& LPIPS$_{\text{inter}}\downarrow$ & 0.384  & 0.542 & \textcolor{blue}{0.282} & \textcolor{red}{0.272} {\small\textbf{(-0.010)}}  \\ \midrule
\multirow{6}{*}{\rotatebox{90}{Set8 50}} & PSNR $\uparrow$ & 21.531 & \textcolor{blue}{23.433} & 23.197 & \textcolor{red}{23.713} {\small\textbf{(+0.516)}} \\
& SSIM $\uparrow$ & 0.501 & 0.482 & \textcolor{blue}{0.594} & \textcolor{red}{0.630} {\small\textbf{(+0.036)}} \\
& LPIPS $\downarrow$ & 0.415 & 0.574 & \textcolor{blue}{0.261} & \textcolor{red}{0.245} {\small\textbf{(-0.016)}} \\
& $E_{\text{warp}}^* \downarrow$ & \textcolor{blue}{0.911} & 1.358 & 1.078 & \textcolor{red}{0.747} {\small\textbf{(-0.331)}} \\
& $E_{\text{inter}} \downarrow$ & \textcolor{blue}{17.217} & 19.845 & 19.732 & \textcolor{red}{16.814} {\small\textbf{(-2.918)}} \\
& LPIPS$_{\text{inter}}\downarrow$ & 0.406 & 0.432 & \textcolor{blue}{0.332} & \textcolor{red}{0.255} {\small\textbf{(-0.077)}} \\ \midrule
\multirow{6}{*}{\rotatebox{90}{Set8 100}} & PSNR $\uparrow$ & 21.226 & 18.198 & \textcolor{blue}{22.519} & \textcolor{red}{22.955} {\small\textbf{(+0.436)}} \\
& SSIM $\uparrow$ & 0.484 & 0.281 & \textcolor{blue}{0.553} & \textcolor{red}{0.591} {\small\textbf{(+0.038)}} \\
& LPIPS $\downarrow$ & 0.472 & 0.733 & \textcolor{blue}{0.338} & \textcolor{red}{0.323} {\small\textbf{(-0.015)}} \\
& $E_{\text{warp}}^* \downarrow$ & \textcolor{blue}{0.918} & 2.229 & 1.13 & \textcolor{red}{0.802} {\small\textbf{(-0.328)}} \\
& $E_{\text{inter}} \downarrow$ & \textcolor{red}{17.367} & 24.661 & 20.18 & \textcolor{blue}{17.444} {\small\textbf{(-2.736)}} \\
& LPIPS$_{\text{inter}}\downarrow$ & 0.421 & 0.619 & \textcolor{blue}{0.372} & \textcolor{red}{0.286} {\small\textbf{(-0.086)}} \\ \midrule
\multirow{6}{*}{\rotatebox{90}{Set8 150}} & PSNR $\uparrow$ & 20.209 & 16.136 & \textcolor{blue}{21.005} & \textcolor{red}{21.418} {\small\textbf{(0.413)}} \\
& SSIM $\uparrow$ & 0.443 & 0.291 & \textcolor{blue}{0.486} & \textcolor{red}{0.544} {\small\textbf{(0.058)}} \\
& LPIPS $\downarrow$ & 0.554 & 0.729 & \textcolor{blue}{0.449} & \textcolor{red}{0.402} {\small\textbf{(-0.047)}} \\
& $E_{\text{warp}}^* \downarrow$ & \textcolor{blue}{0.972} & 4.279 & 1.207 & \textcolor{red}{0.832} {\small\textbf{(-0.375)}} \\
& $E_{\text{inter}} \downarrow$ & \textcolor{blue}{17.872} & 22.343 & 20.729 & \textcolor{red}{17.616} {\small\textbf{(-3.113)}} \\
& LPIPS$_{\text{inter}}\downarrow$ & 0.470 & 0.646 & \textcolor{blue}{0.450} & \textcolor{red}{0.331} {\small\textbf{(-0.119)}} \\
\bottomrule
\end{tabular}
}
\end{table}
\begin{figure}[t]
\centering
{
\includegraphics[width=\columnwidth]{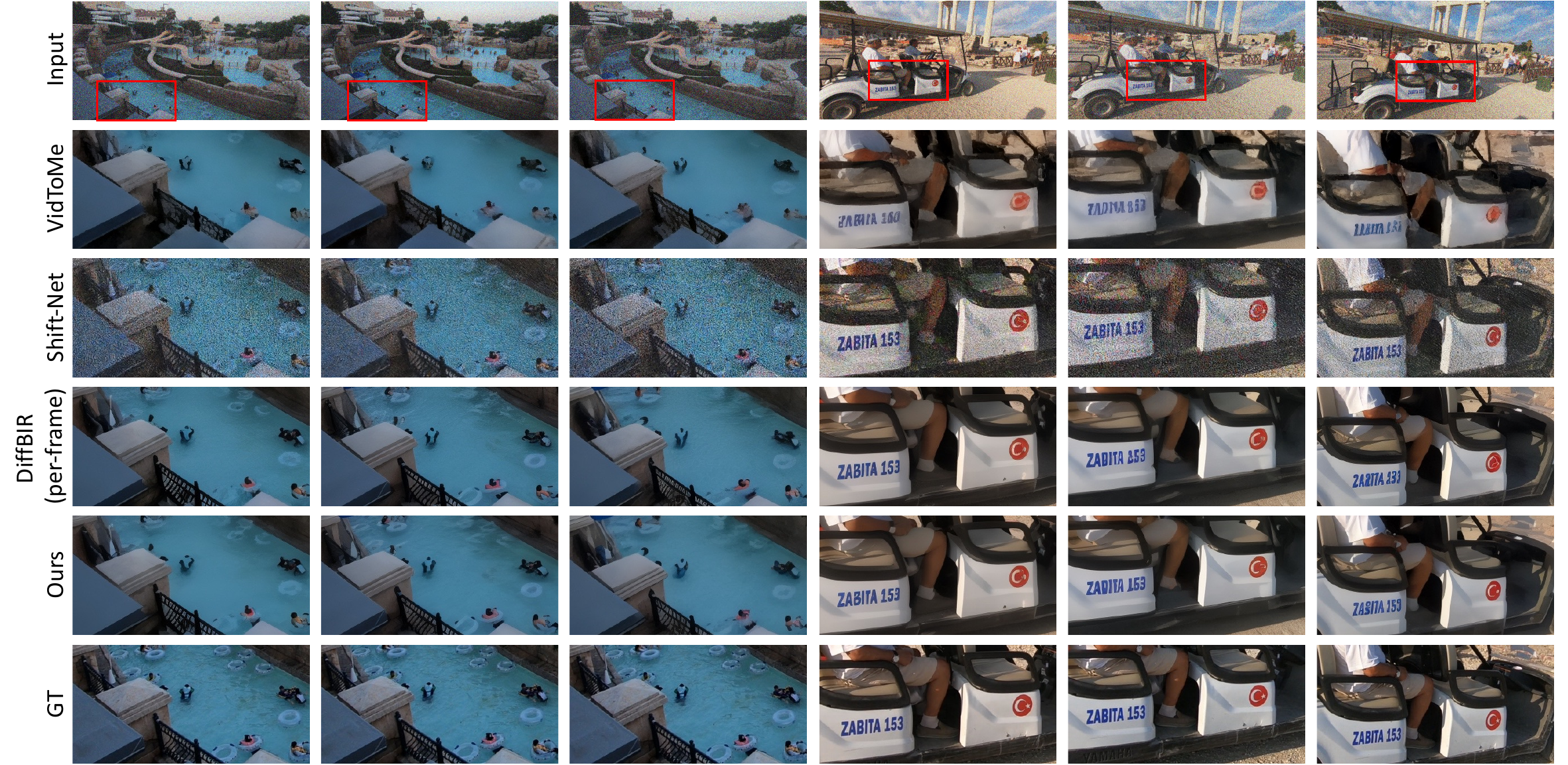}
}
\vspace{-7mm}
\caption{\textbf{
Video denoising comparisons on the REDS30~\citep{nah2019ntire} dataset.
} Our method effectively denoises and generates detailed results while maintaining temporal coherence.
}
\label{fig:visual_Denoise}
\end{figure}

\noindent {\bf Video denoising.}
In denoising experiments in~\cref{tab:denoise}, we observe that while regression models can perform adequately with sufficient batch sizes, our method consistently achieves superior perceptual quality (LPIPS) and maintains this advantage even under severe degradation. As shown in \cref{fig:visual_Denoise}, Shift-Net~\citep{Li_2023_CVPR} struggles with out-of-distribution noise levels, and VidToMe produces temporally consistent but detail-deficient results. Per-frame DiffBIR generates high-quality frames but suffers from temporal inconsistencies, particularly noticeable in facial features and moving objects. Our method successfully balances detail preservation with temporal consistency.

\begin{figure}[]
\centering
\footnotesize
\setlength{\tabcolsep}{1pt}
\renewcommand{\arraystretch}{1}
\resizebox{\columnwidth}{!}{%
\begin{tabular}{cccccc}
\raisebox{2.2\normalbaselineskip}[0pt][0pt]{\rotatebox[origin=c]{90}{Input}} &
\includegraphics[width=0.2\textwidth]{figures/depth/input_goat_00043.jpg} &
\includegraphics[width=0.2\textwidth]{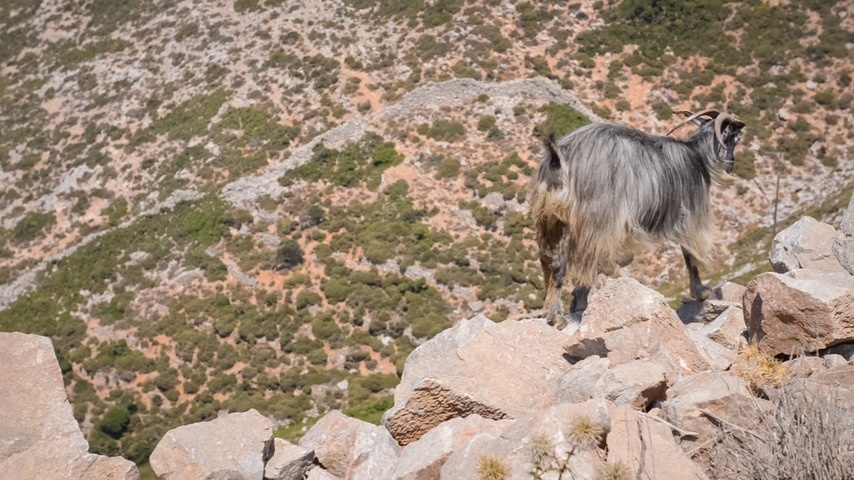} &
\includegraphics[width=0.2\textwidth]{figures/depth/input_goat_00045.jpg} &
\includegraphics[width=0.2\textwidth]{figures/depth/input_goat_00046.jpg} &
\includegraphics[width=0.2\textwidth]{figures/depth/input_goat_00047.jpg} \\
\raisebox{2.2\normalbaselineskip}[0pt][0pt]{\rotatebox[origin=c]{90}{Marigold}} &
\includegraphics[width=0.2\textwidth]{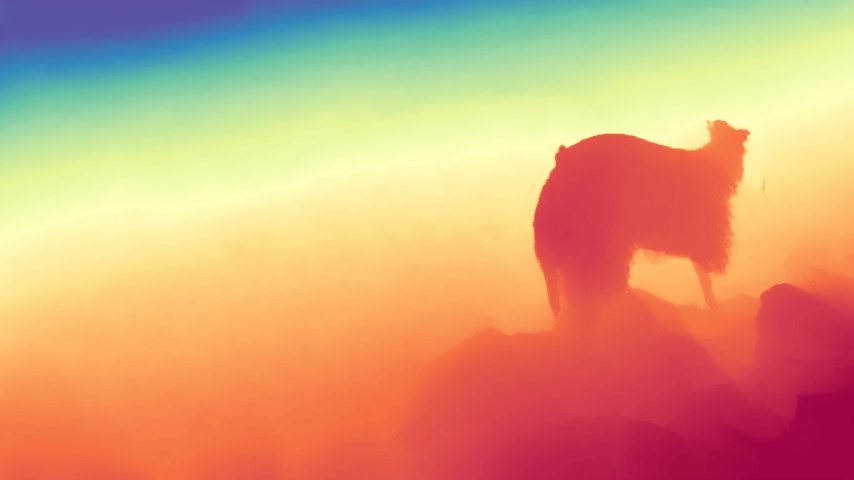} &
\includegraphics[width=0.2\textwidth]{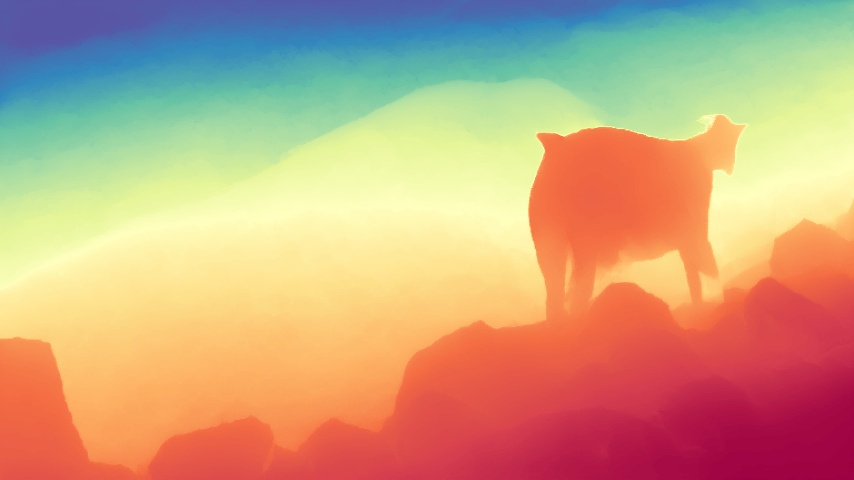} &
\includegraphics[width=0.2\textwidth]{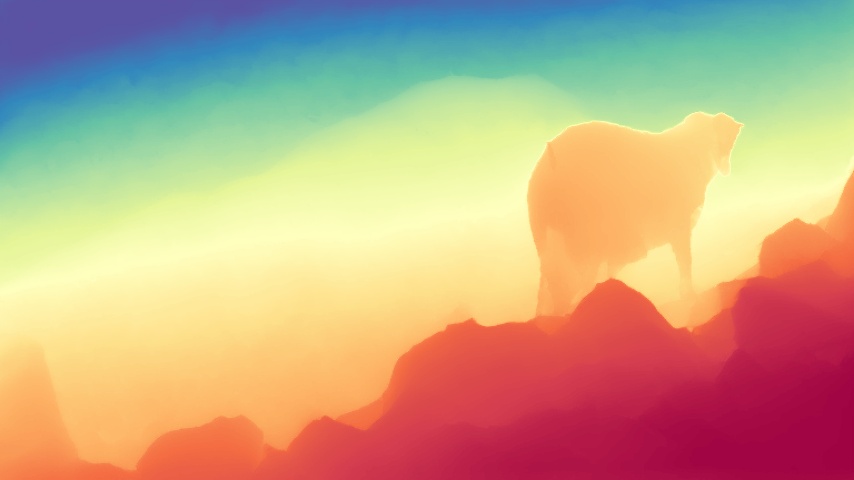} &
\includegraphics[width=0.2\textwidth]{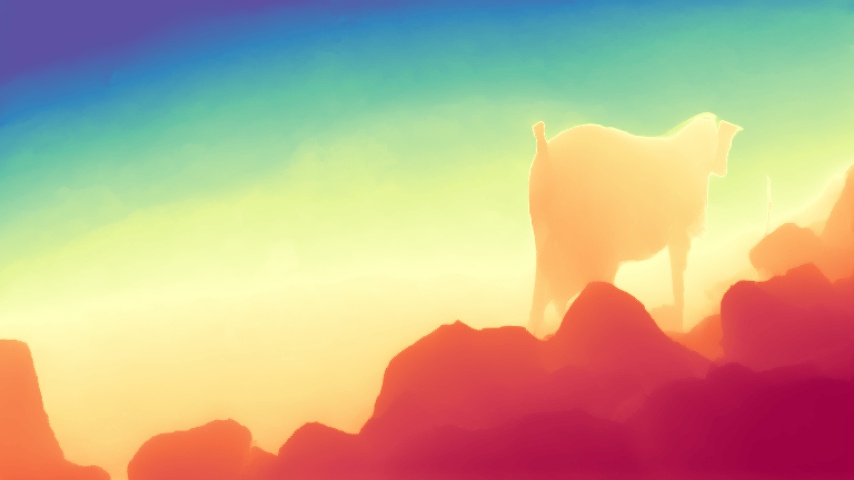} &
\includegraphics[width=0.2\textwidth]{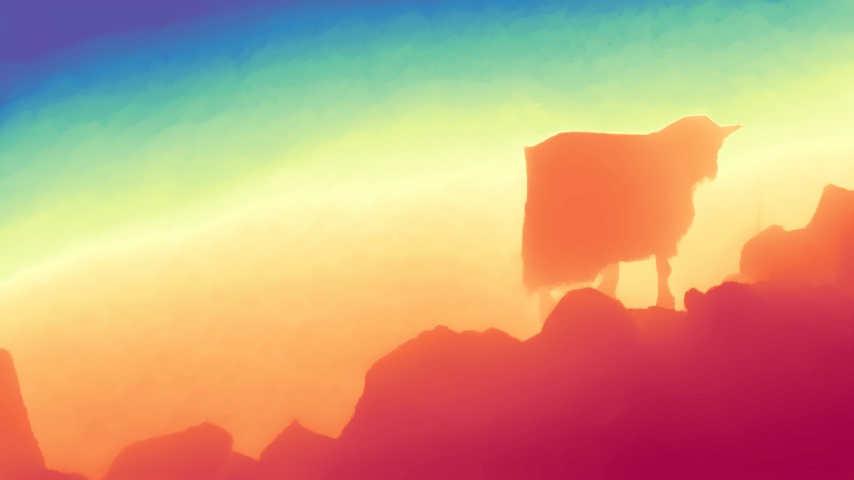} \\
\raisebox{2.2\normalbaselineskip}[0pt][0pt]{\rotatebox[origin=c]{90}{Ours}} &
\includegraphics[width=0.2\textwidth]{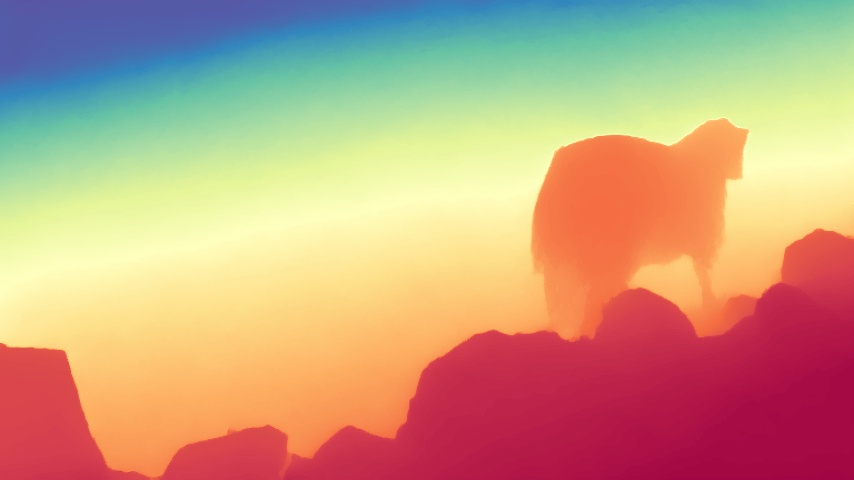} &
\includegraphics[width=0.2\textwidth]{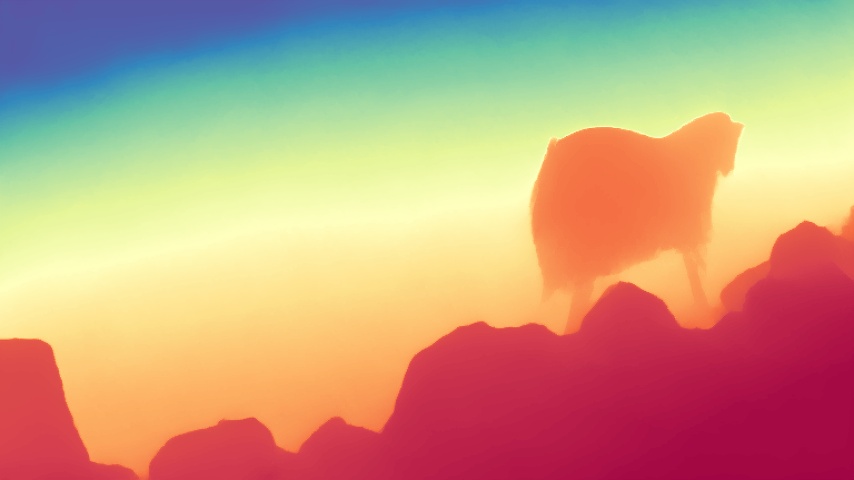} &
\includegraphics[width=0.2\textwidth]{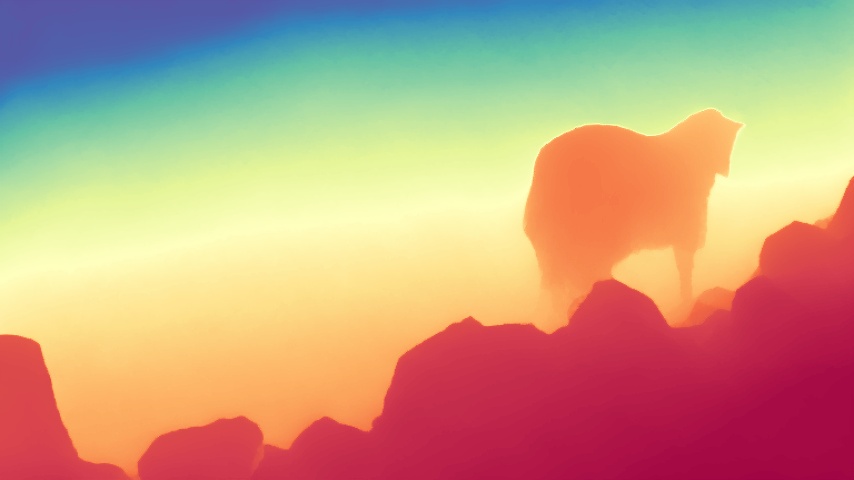} &
\includegraphics[width=0.2\textwidth]{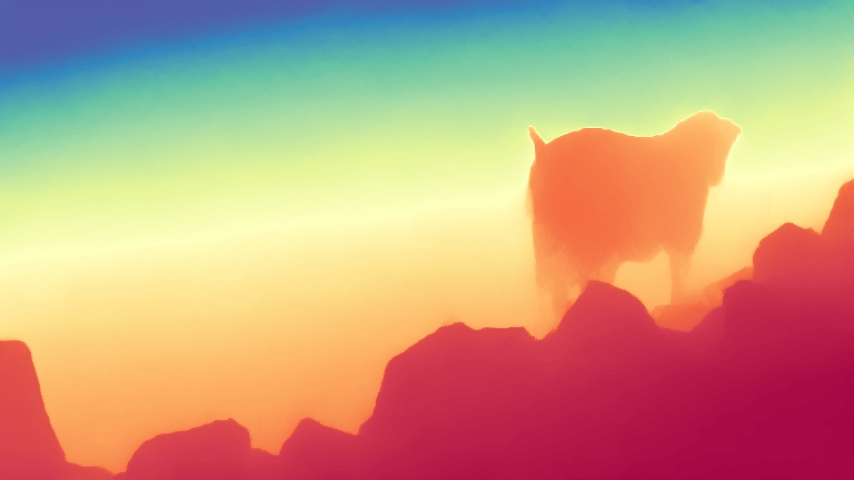} &
\includegraphics[width=0.2\textwidth]{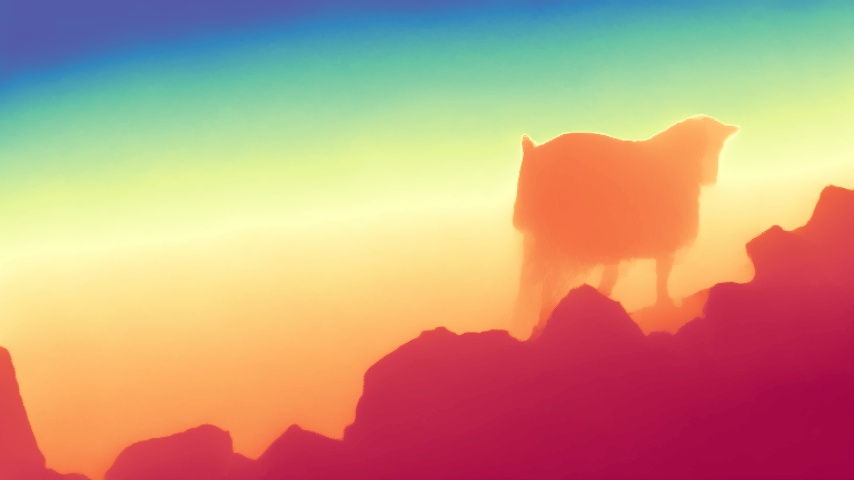} \\
\end{tabular}%
}
\vspace{-3mm}
\caption{\textbf{Applying our techniques to consistent video depth.} Integrating our proposed framework into Marigold~\citep{ke2024repurposing} helps improve the temporal consistency of video depth estimation.
}
\label{fig:depth_visual}
\end{figure}

\noindent {\bf Generalization to other tasks.}
To demonstrate the broad applicability of our framework, we integrate it with Marigold~\citep{ke2024repurposing}, a state-of-the-art monocular depth estimator based on latent diffusion. Results in \cref{fig:depth_visual} show significant improvements in temporal consistency of depth estimates while maintaining accuracy. This successful adaptation to depth estimation, alongside our results in super-resolution and denoising, highlights the versatility of our approach. As the field advances and more powerful image models emerge, our framework's zero-shot nature allows immediate leverage of these improvements across video restoration tasks.

\begin{table}[t]
\centering
\caption{\textbf{Ablation studies for 8$\times$ VSR with different correspondence matching methods on DAVIS~\citep{7780454} test sets.} 
}
\vspace{-3mm}
\label{tab:ablation_matching}
\resizebox{\columnwidth}{!}{
\setlength\columnsep{4pt} 
\begin{tabular}{llc|ccc}
\toprule
\makecell{Down\\blocks} & \makecell{Up\\blocks} & \makecell{Spatial-\\aware} & LPIPS $\downarrow$ & $E_{\text{warp}}^* \downarrow$  & LPIPS$_{\text{inter}}\downarrow$ \\
\midrule
Flow & Flow & -- & 0.518 & 1.214 & 0.563 \\
Cos & Cos & -- & 0.390 & 0.736 & 0.350 \\
Cos & Flow & -- & 0.507 & 1.049 & 0.545 \\
Flow & Cos & -- & 0.375 & \best{0.677} & 0.347 \\
Flow & Cos & \checkmark &  \best{0.367} & 0.699 & \best{0.333} \\
\bottomrule
\end{tabular}
}
\end{table}

\begin{table}[t]
\centering
\caption{\textbf{Ablation studies for 8$\times$ VSR with the proposed components applied at different stages of the denoising process on DAVIS~\citep{7780454} test sets.} We apply our two proposed components, hierarchical latent warping (HLW) and hybrid spatial-aware token merging (HS-ToMe), at the early, mid, and late denoising stages.
}
\vspace{-3mm}
\label{tab:ablation_hlw}
\resizebox{\columnwidth}{!}{
\setlength\columnsep{4pt} 
\begin{tabular}{ccccccc|ccc}
\toprule
\multicolumn{3}{c}{HLW (Sec.~\ref{subsec:3_3})} & & \multicolumn{3}{c|}{HS-ToMe (Sec.~\ref{subsec:3_4})} \\ \cmidrule{1-3} \cmidrule{5-7}
\centering Early & \centering Mid & \centering Late & & \centering Early & \centering Mid & \centering Late & LPIPS $\downarrow$ & $E_{\text{warp}}^* \downarrow$  & LPIPS$_{\text{inter}}\downarrow$ \\
\midrule
\centering -- & \centering -- & \centering -- & & \centering -- & \centering -- & \centering -- & \textbf{0.362} & 0.964 & 0.372 \\
\centering \checkmark & \centering -- & \centering -- & & \centering \checkmark & \centering -- & \centering -- & 0.368 & 0.887 & 0.369 \\
\centering \checkmark & \centering \checkmark & \centering -- & & \centering \checkmark & \centering \checkmark & \centering \checkmark & 0.43 & 0.804 & 0.383 \\
\centering \checkmark & \centering \checkmark & \centering \checkmark & & \centering \checkmark & \centering \checkmark & \centering \checkmark & 0.411 & 0.704 & 0.339 \\
\centering \checkmark & \centering -- & \centering -- & & \centering \checkmark & \centering \checkmark & \centering \checkmark & 0.367 & \textbf{0.699} & \textbf{0.333} \\
\bottomrule
\end{tabular}
}
\end{table}

\begin{table}[t]
\centering
\small
\caption{\textbf{Ablation studies on the merging ratio $r$.}
}
\vspace{-3mm}
\label{tab:ablation_merging_ratio}
\setlength\columnsep{4pt} 
\begin{tabular}{l|ccc}
\toprule
Merging ratio $r$ & PSNR $\uparrow$ & SSIM $\uparrow$  & LPIPS $\downarrow$ \\
\midrule
0.3 & 22.814 & 0.483 & 0.358 \\
0.6 & 23.308 & 0.522 & 0.477 \\
0.9 & 23.169 & 0.518 & 0.478 \\
0.6 $\rightarrow$ 0 & 23.143 & 0.507 & 0.403 \\
0.9 $\rightarrow$ 0 (Ours) & 23.302 & 0.518 & 0.428 \\
\bottomrule
\end{tabular}
\end{table}

\subsection{Ablation Study}


We conduct extensive ablation studies to validate our design choices and analyze the impact of different components on the final performance. These experiments not only demonstrate the effectiveness of our approach but also provide insights into the interaction between different components.

\noindent {\bf Correspondence mechanism selection.}
We first examine different strategies for identifying token correspondences across frames. As shown in \cref{tab:ablation_matching}, we systematically evaluate combinations of optical flow and cosine similarity at different stages of the UNet architecture. Our hybrid approach—using optical flow in downsample blocks and cosine similarity in upsample blocks—achieves optimal results across all metrics. This validates our insight that different correspondence mechanisms are more effective at different network depths. The addition of spatial awareness further improves performance by preventing mismatches in textureless regions and ensuring locally coherent correspondence matches.

\noindent {\bf Component scheduling analysis.}
We analyze the effectiveness of our two key components—hierarchical latent warping (HLW) and hybrid spatial-aware token merging (HS-ToMe)—when applied at different stages of the denoising process. \cref{tab:ablation_hlw} shows that applying latent warping during mid or late denoising stages significantly degrades performance. This confirms our hypothesis that latent manipulation is most effective in early stages when establishing coarse structure, while token-level operations are crucial throughout the process for maintaining fine details and temporal consistency.

A particularly interesting finding is that attempting to enforce consistency in later stages through latent warping can actually harm the restoration quality, likely due to the increasing semantic richness of the latent space. This insight guided our design choice to transition from latent-based to token-based consistency enforcement as denoising progresses. Extensive temporal profile comparisons and additional ablation results are provided in the supplementary materials.

\noindent {\bf Token merging ratio $r$.}
The effectiveness of our method is significantly influenced by the token merging ratio r, which controls the balance between temporal consistency and detail preservation. We conduct extensive ablation studies on this hyperparameter, comparing our annealing strategy (reducing r from 0.9 to 0) against fixed ratios. \cref{tab:ablation_merging_ratio} show that higher fixed ratios (0.9, 0.6) tend to improve fidelity metrics (PSNR/SSIM) by enforcing stronger temporal consistency, but at the cost of perceptual quality (higher LPIPS) due to over-smoothing. Conversely, lower fixed ratios (0.3) preserve more details but sacrifice temporal coherence. Our annealing strategy achieves the best overall performance (PSNR: 23.302, SSIM: 0.518, LPIPS: 0.428) by adaptively reducing the merging ratio throughout the denoising process, effectively balancing the trade-off between temporal consistency and detail preservation.
\section{Conclusion}
We have presented DiffIR2VR-Zero, a novel framework enabling zero-shot video restoration using pre-trained image diffusion models without additional training. Our approach combines hierarchical latent warping with hybrid flow-guided token merging to maintain temporal consistency while preserving high-quality restoration capabilities. Extensive experiments demonstrate state-of-the-art performance across various restoration tasks and remarkable robustness to severe degradations.

\noindent {\bf Limitations.}
Our framework faces two main limitations: flickering artifacts in dynamic scenes due to LDM decoder sensitivity, and reduced performance under extreme degradation scenarios. Future work will focus on stabilizing decoder output and enhancing degradation handling capabilities. As more powerful image diffusion models emerge, our framework's zero-shot nature will allow immediate leverage of these advances for improved video restoration.

{\small
\bibliographystyle{ieeenat_fullname}
\bibliography{11_references}
}

\ifarxiv \clearpage \appendix \section{Appendix Section}
\label{sec:appendix_section}
In this supplementary material, we first provide additional details on the testing datasets and evaluation metrics. Then, we report the computational complexity and inference time comparison. Subsequently, we present more visual comparisons of various methods.

\subsection{Ablation Studies on Correspondences Identified by Cosine Similarity}
\cref{fig:correspondence_pad_fig_v2} The figure shows the correspondences at denoising step 40 for three scenarios: without spatial awareness and padding removal, without spatial awareness, and with both spatial awareness and padding removal (ours). It is evident that padding values significantly affect the matching quality. However, even after removing padding, many mismatched diagonal lines remain, leading to blurry results. In contrast, our method effectively finds accurate correspondences by leveraging spatial information from the video. 

\begin{figure}[t]
    \centering
    \begin{subfigure}{\columnwidth}
        \centering
        \includegraphics[width=\textwidth]{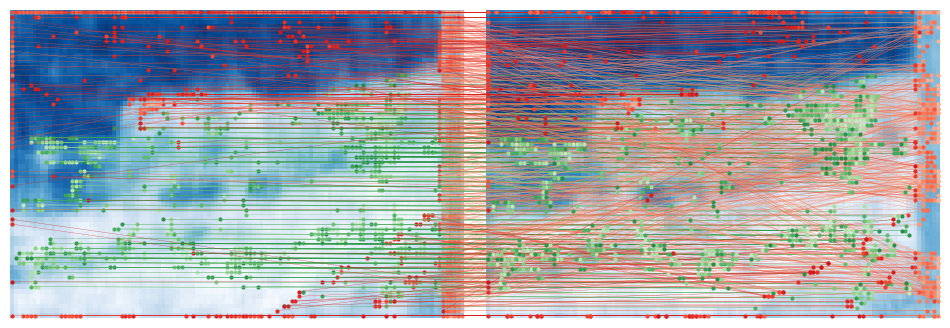}
        \caption{Correspondences without spatial-awareness and padding removal}
    \end{subfigure}
    

    \begin{subfigure}{\columnwidth}
        \centering
        \includegraphics[width=\textwidth]{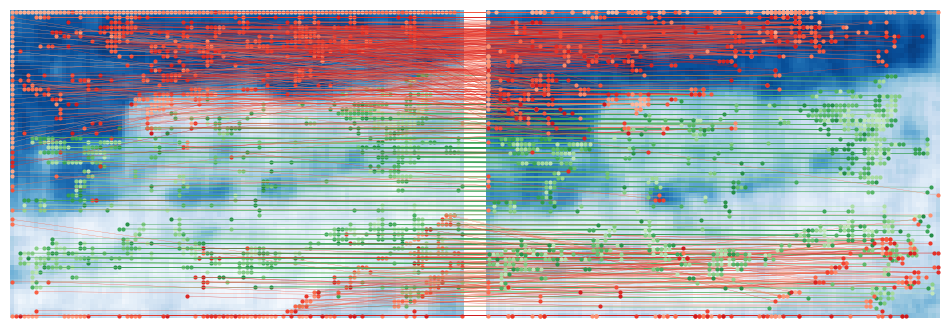}
        \caption{Correspondences without spatial-awareness}
    \end{subfigure}
    

    \begin{subfigure}{\columnwidth}
        \centering
        \includegraphics[width=\textwidth]{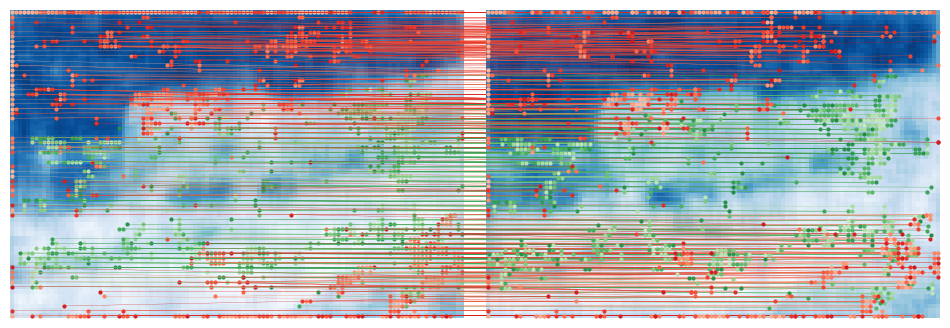}
        \caption{Correspondences with spatial-awareness and padding removal (ours)}
    \end{subfigure}
    
    \caption{\textbf{Correspondences at denoising step 40 for different settings.}}
    \label{fig:correspondence_pad_fig_v2}
\end{figure}



\subsection{Severe Degradation Scenarios}
Our balanced approach proves particularly effective in severe degradation scenarios. For instance, in 8$\times$ super-resolution tasks, our method not only avoids artifacts but can even improve visual quality compared to per-frame approaches (\cref{fig:rebuttal_extreme}). Additionally, in the 4$\times$ video face super-resolution dataset~\citep{chen2024towards}, our results contain more details compared to FMA-Net and are temporally more consistent than per-frame method DiffBIR as shown in \cref{fig:visual_face_SR}.
This underscores the effectiveness of our ratio annealing technique in addressing the over-smoothing tendency while maintaining the benefits of our token merging approach. Additional comparisons on video super-resolution can be found at \cref{fig:visual_SR_appendix_2} and \cref{fig:visual_SR_appendix}.

\begin{figure}[t]
    \centering
    \includegraphics[width=\columnwidth]{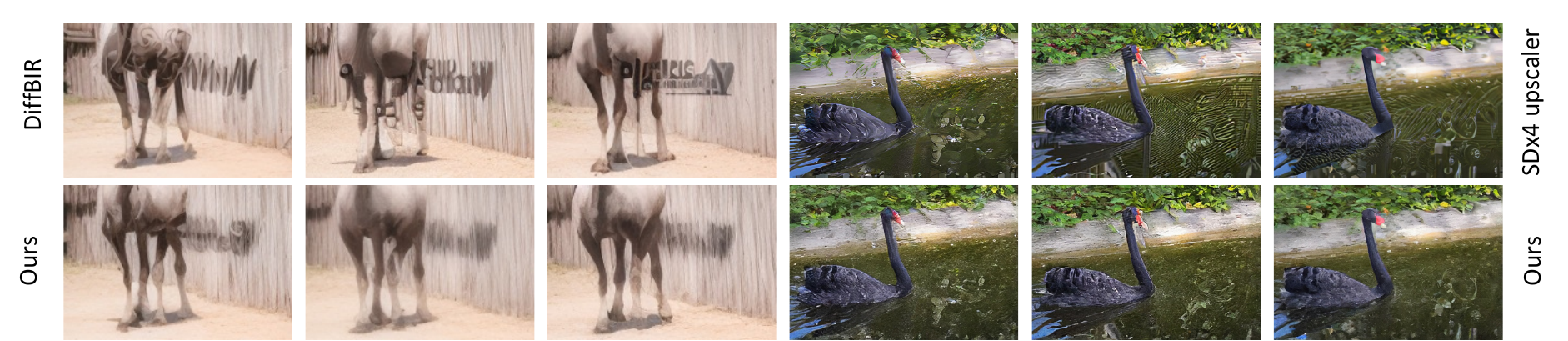}
    \caption{\textbf{Applying our method on DiffBIR and SD $\times$4 upscaler for 8$\times$SR task.} In this case of severe degradation, our method avoids artifacts and outperforms per-frame inference in terms of visual quality.}
    \label{fig:rebuttal_extreme}
\end{figure}

\begin{figure}[t]
\centering
{
\includegraphics[width=\columnwidth]{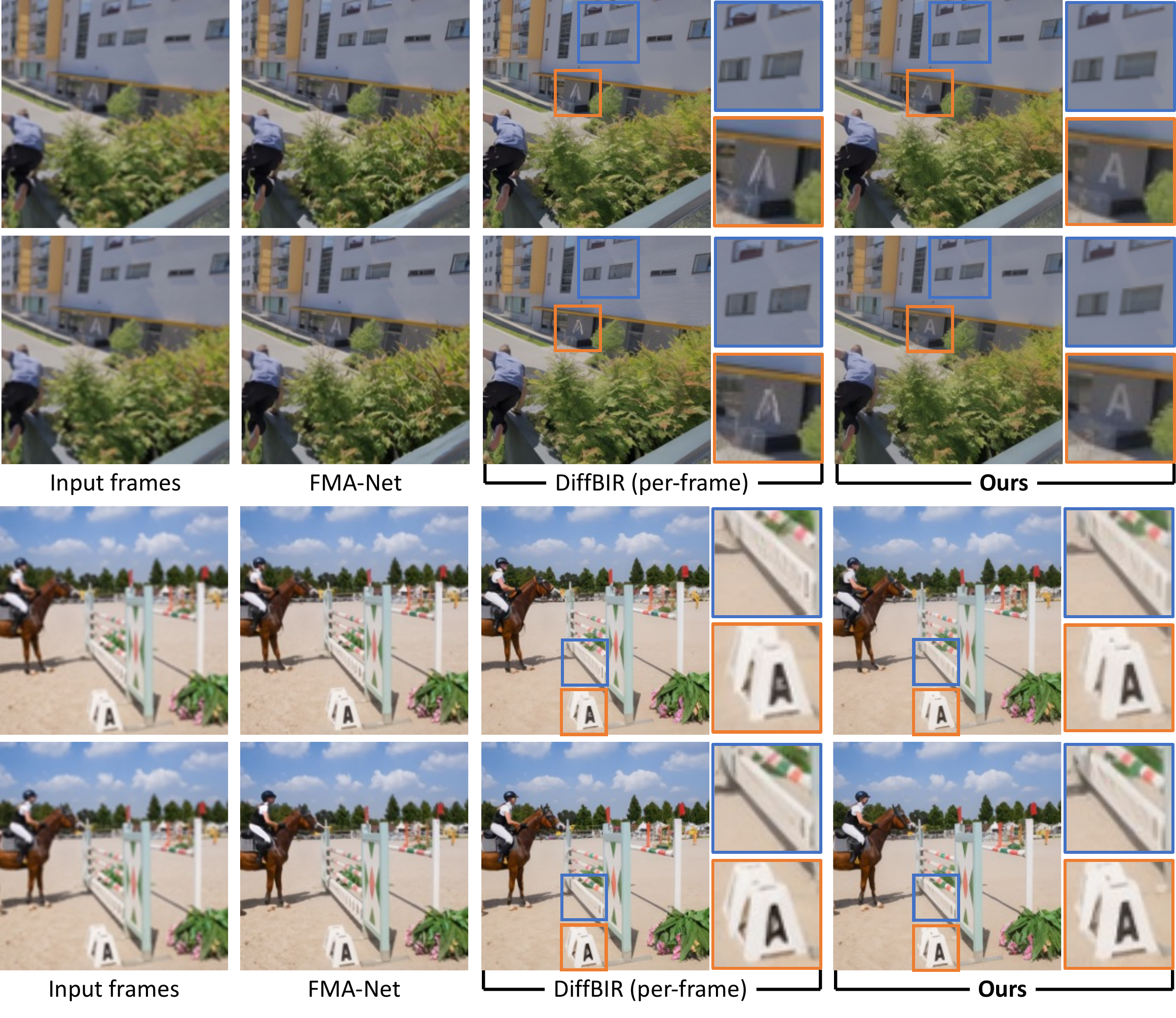}
}
\caption{\textbf{Additional qualitative comparisons on 4$\times$ video super-resolution.} In the zoomed-in patches, our method produces clearer and more consistent results.}
\label{fig:visual_SR_appendix_2}
\end{figure}

\begin{figure}[t]
\centering
{
\includegraphics[width=\columnwidth]{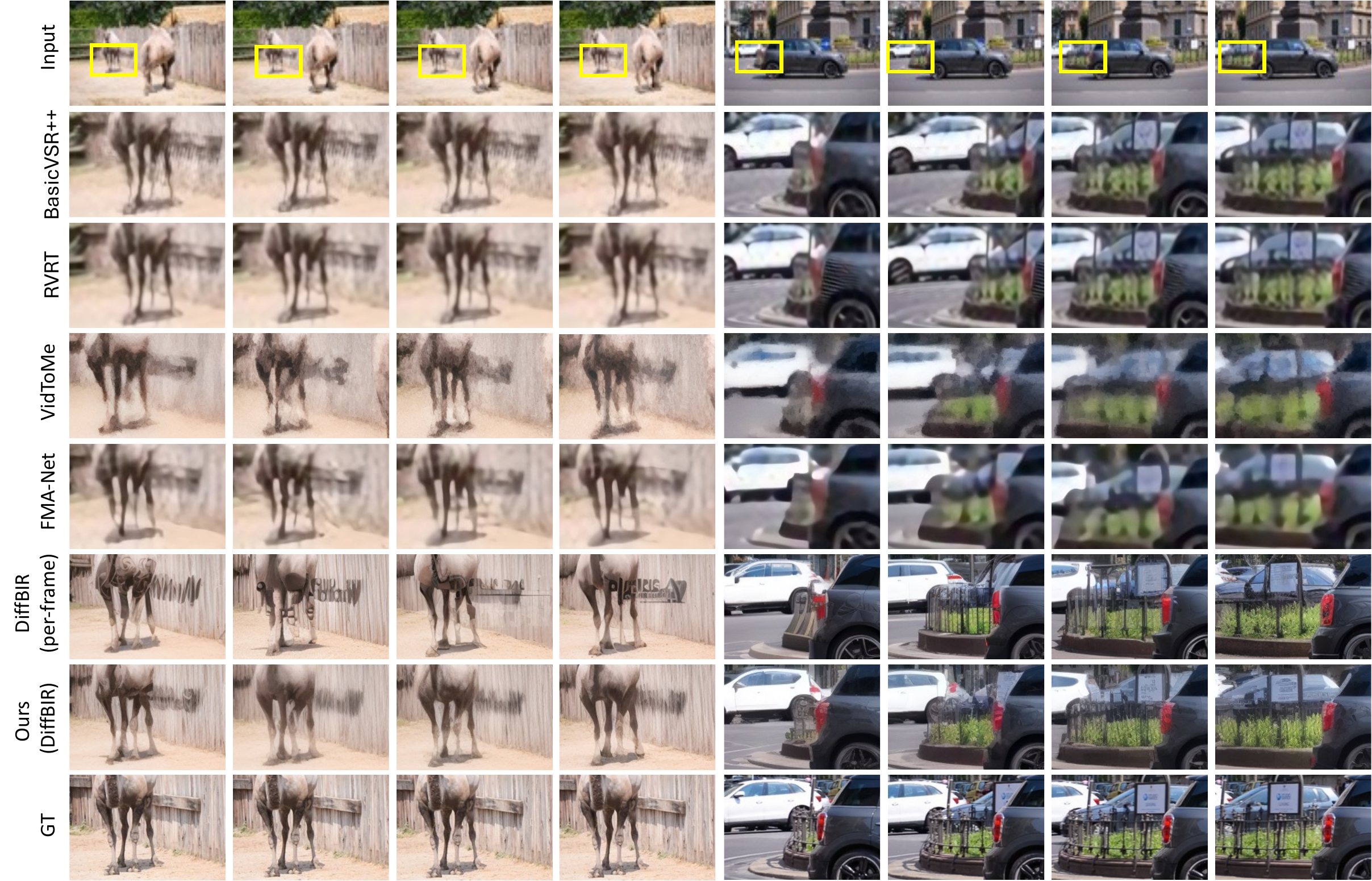}
}
\caption{\textbf{Additional qualitative comparisons on 8$\times$ video super-resolution.} 
As shown in the first row, the low-quality input lacks almost all details. In the zoomed-in patches, our method produces clearer and more consistent results.
}
\label{fig:visual_SR_appendix}
\end{figure}

\begin{figure}[t]
\centering
{
\includegraphics[width=\columnwidth]{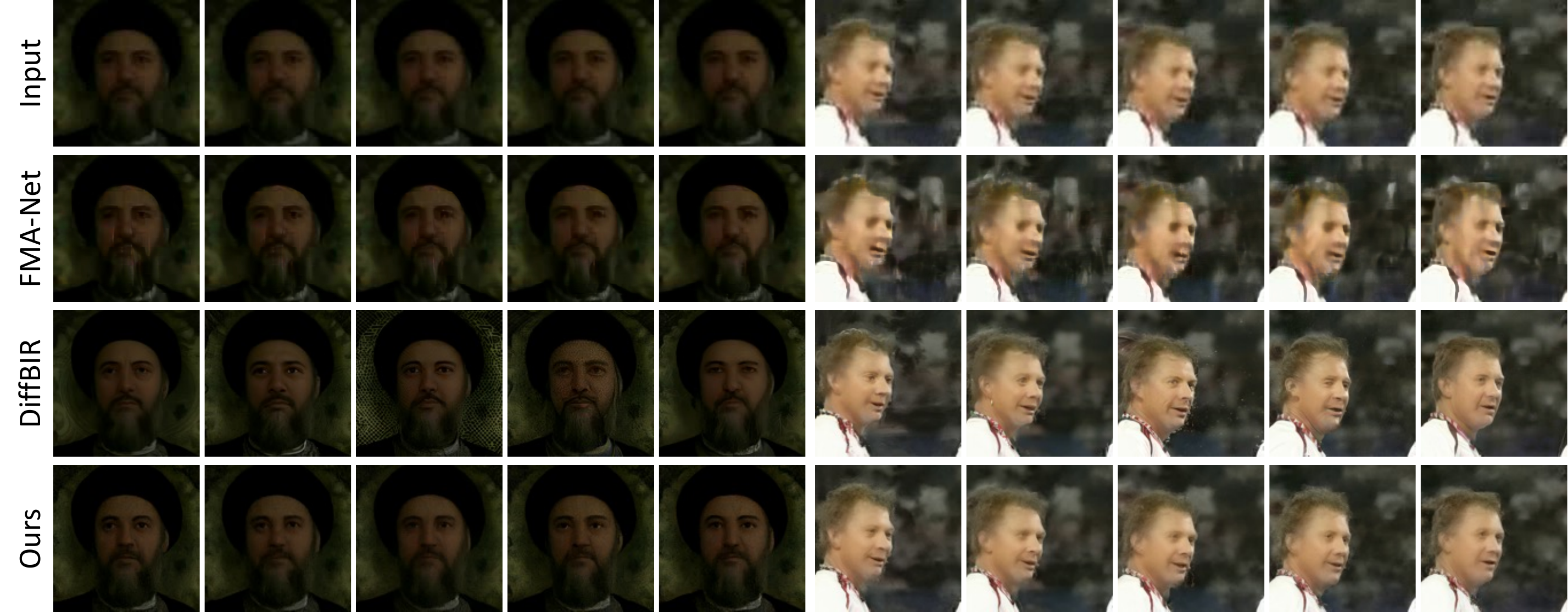}
}

\caption{\textbf{Additional qualitative comparisons on 4$\times$ video face super-resolution.}
}
\label{fig:visual_face_SR}
\end{figure}

\paragraph{Other Video Tasks: Consistent Video Depth.}
Our zero-shot framework is applicable to any pre-trained image-based diffusion models and could improve the predicted video consistency. Therefore, we integrate our proposed zero-shot framework into a state-of-the-art latent diffusion-based monocular depth estimator: Marigold~\citep{ke2024repurposing}. \cref{fig:depth_visual_suppl} shows that integrating our proposed framework into Marigold helps improve the temporal consistency of video depth estimation.

\begin{figure*}[t]
\centering
\footnotesize
\setlength{\tabcolsep}{1pt}
\renewcommand{\arraystretch}{1}
\resizebox{0.9\textwidth}{!}{%
\begin{tabular}{cccccc}
\toprule
\raisebox{2.2\normalbaselineskip}[0pt][0pt]{\rotatebox[origin=c]{90}{Input}} &
\includegraphics[width=0.2\textwidth]{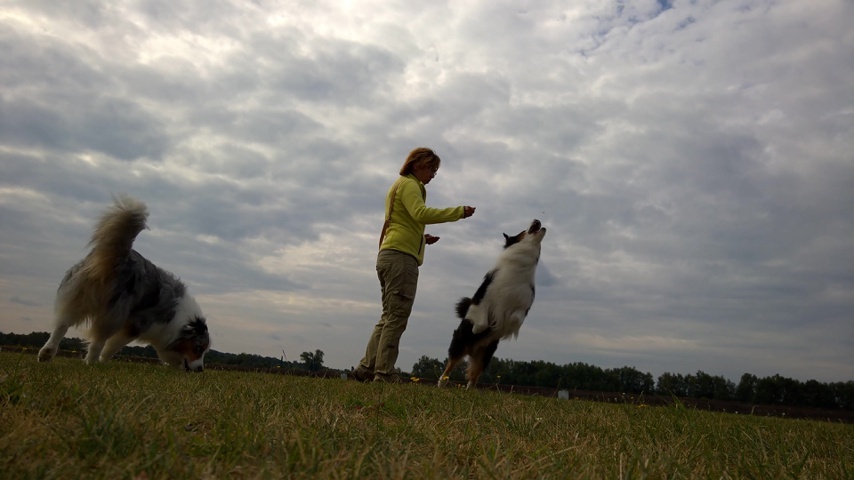} & 
\includegraphics[width=0.2\textwidth]{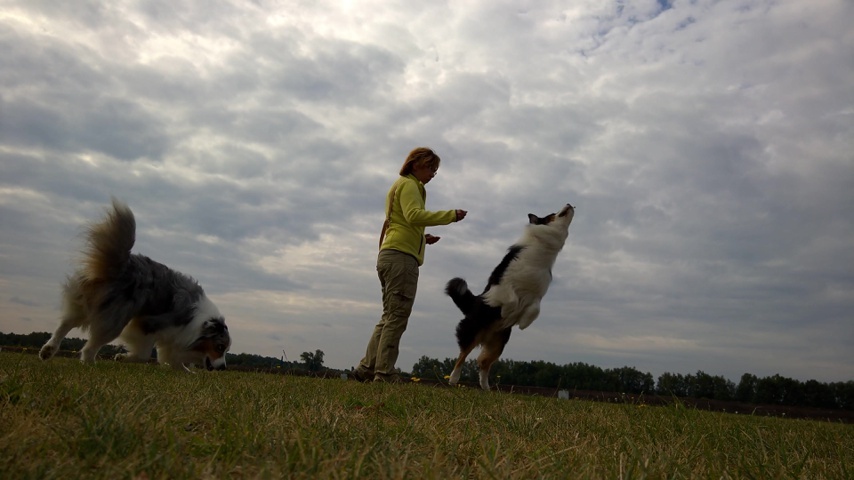} & 
\includegraphics[width=0.2\textwidth]{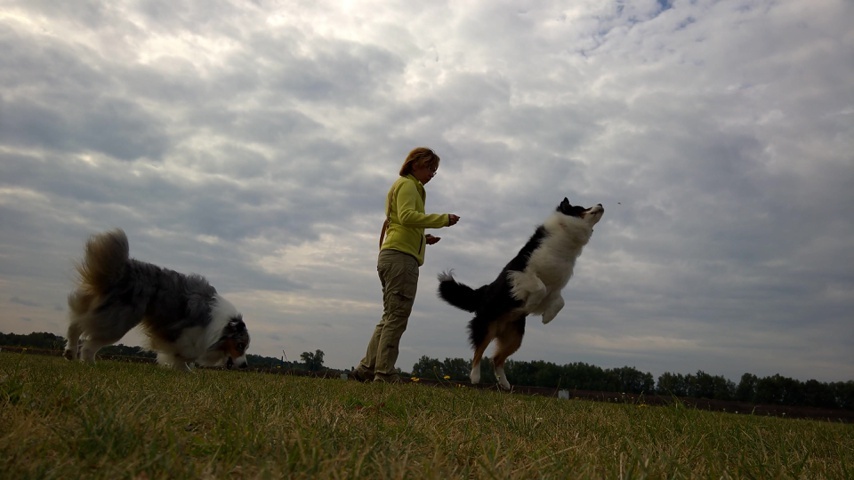} & 
\includegraphics[width=0.2\textwidth]{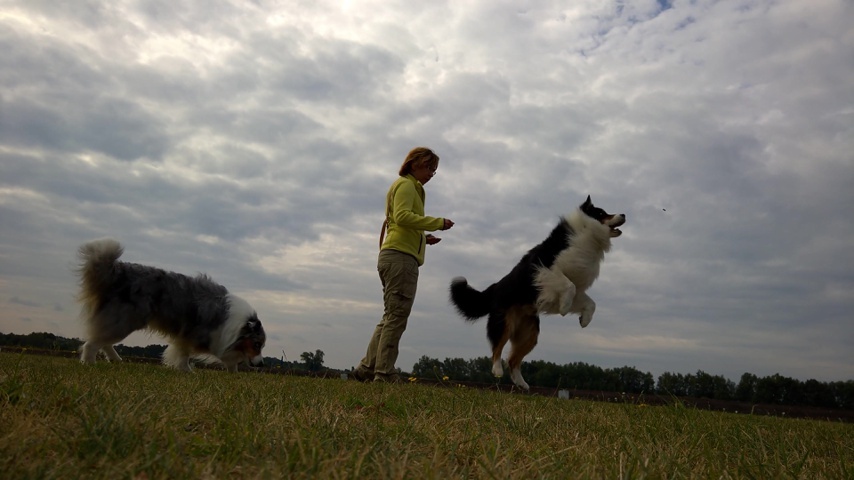} & 
\includegraphics[width=0.2\textwidth]{figures/depth/input_dogs-jump_00035.jpg} \\
\raisebox{2.2\normalbaselineskip}[0pt][0pt]{\rotatebox[origin=c]{90}{Marigold}} &
\includegraphics[width=0.2\textwidth]{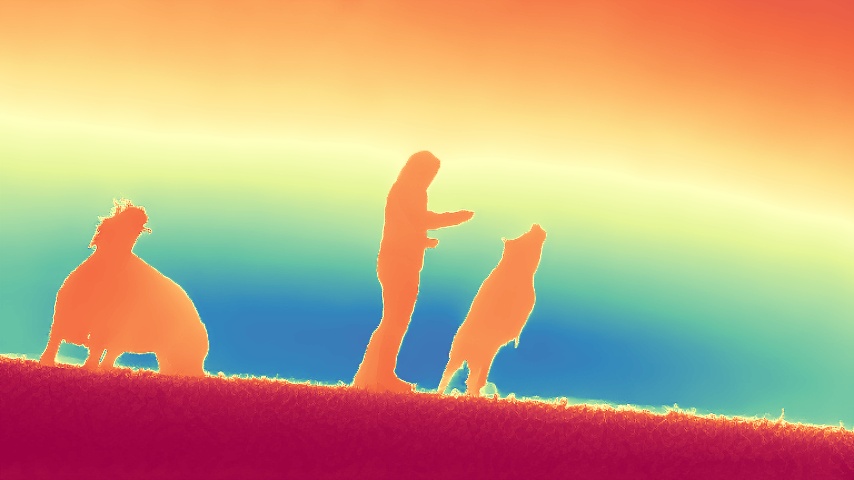} & 
\includegraphics[width=0.2\textwidth]{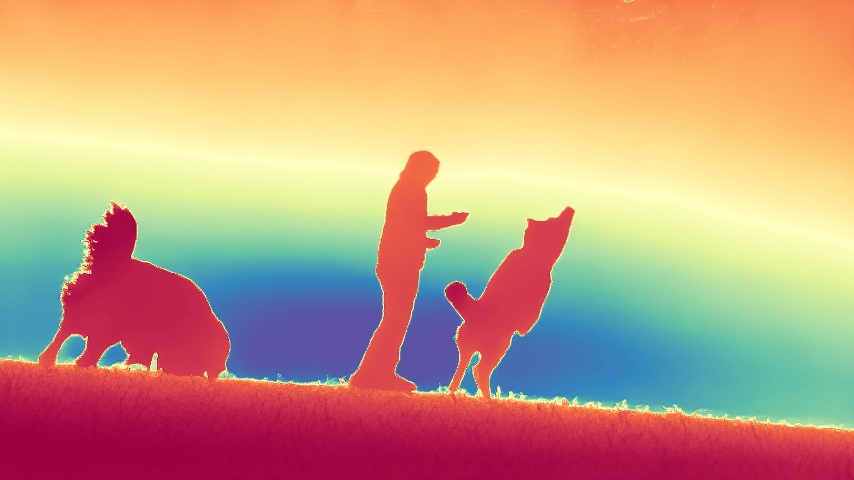} & 
\includegraphics[width=0.2\textwidth]{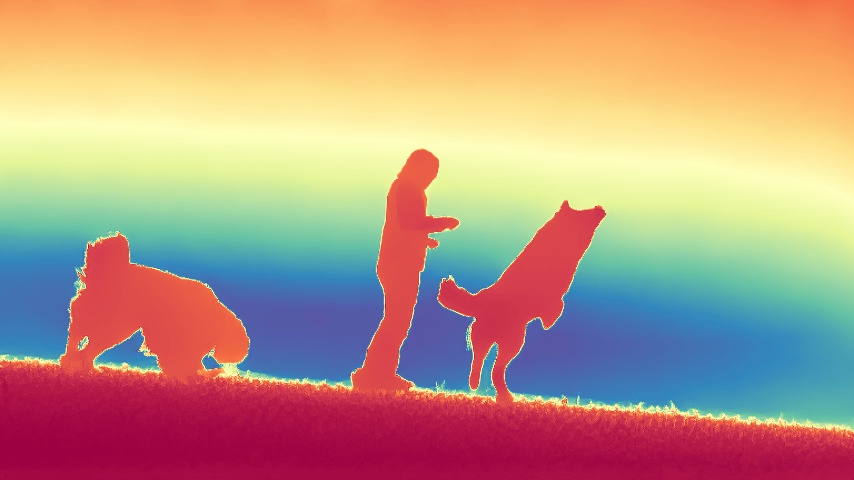} & 
\includegraphics[width=0.2\textwidth]{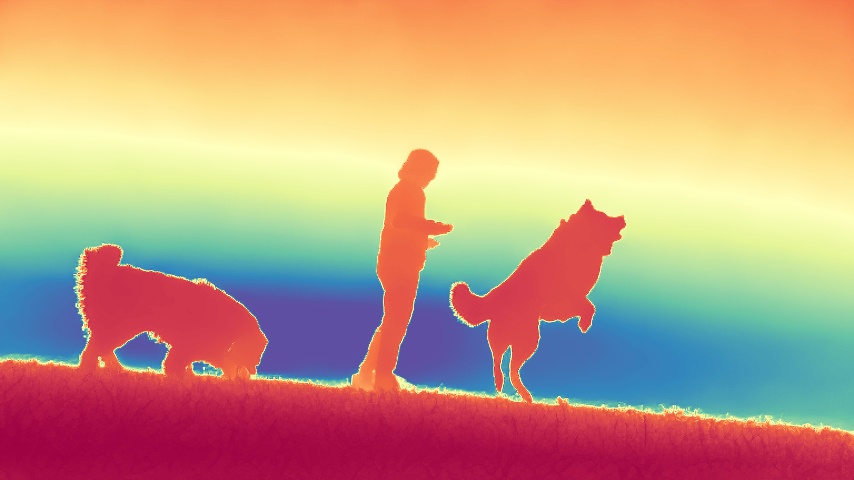} & 
\includegraphics[width=0.2\textwidth]{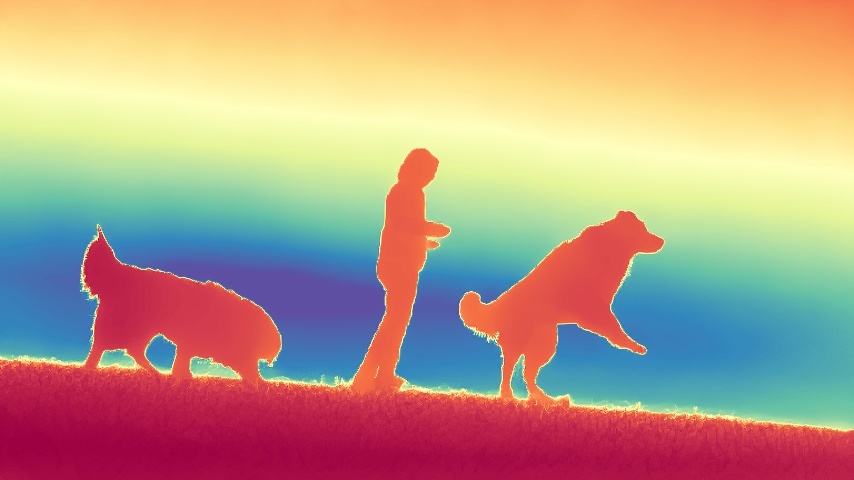} \\
\raisebox{2.2\normalbaselineskip}[0pt][0pt]{\rotatebox[origin=c]{90}{Ours}} &
\includegraphics[width=0.2\textwidth]{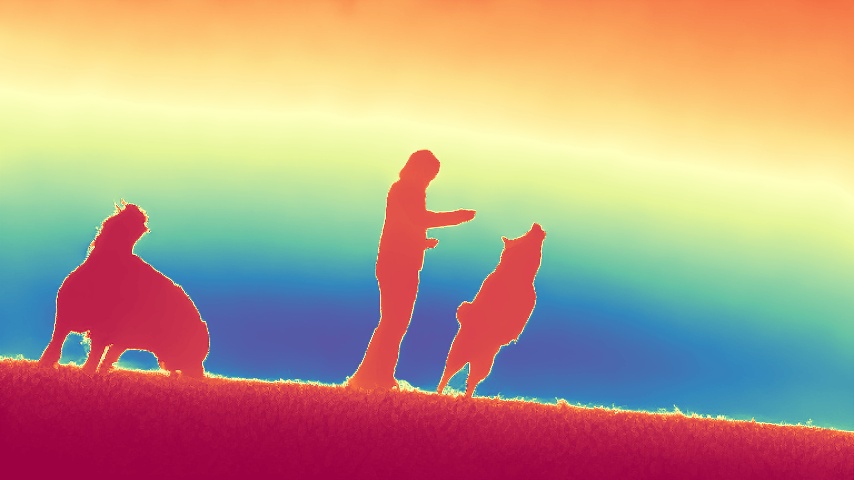} & 
\includegraphics[width=0.2\textwidth]{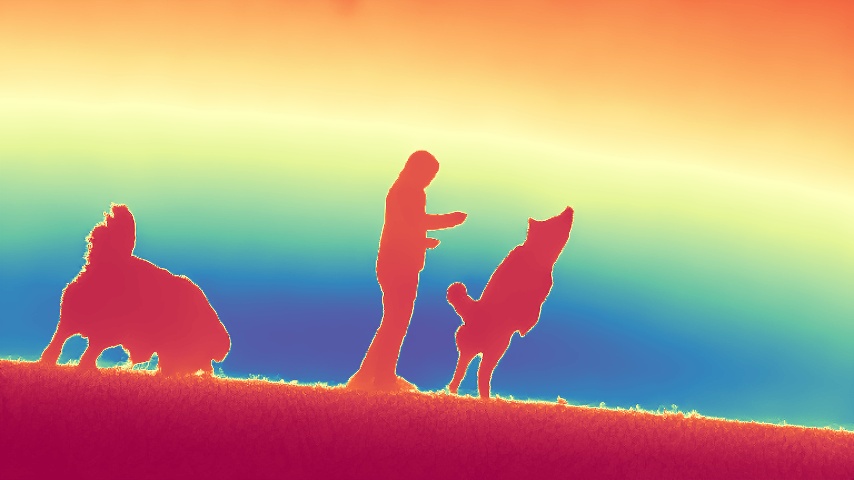} & 
\includegraphics[width=0.2\textwidth]{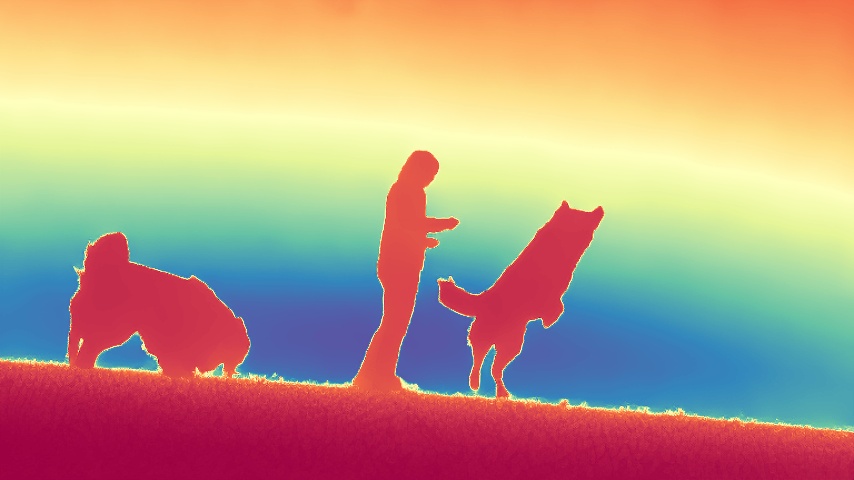} & 
\includegraphics[width=0.2\textwidth]{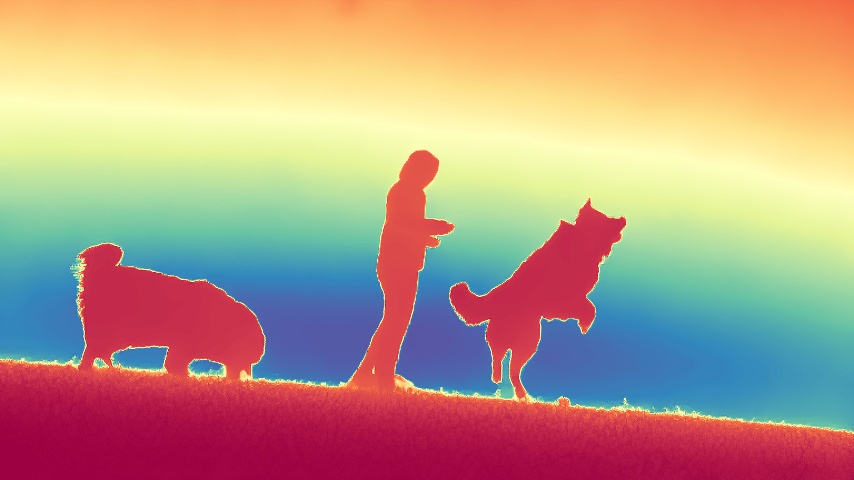} & 
\includegraphics[width=0.2\textwidth]{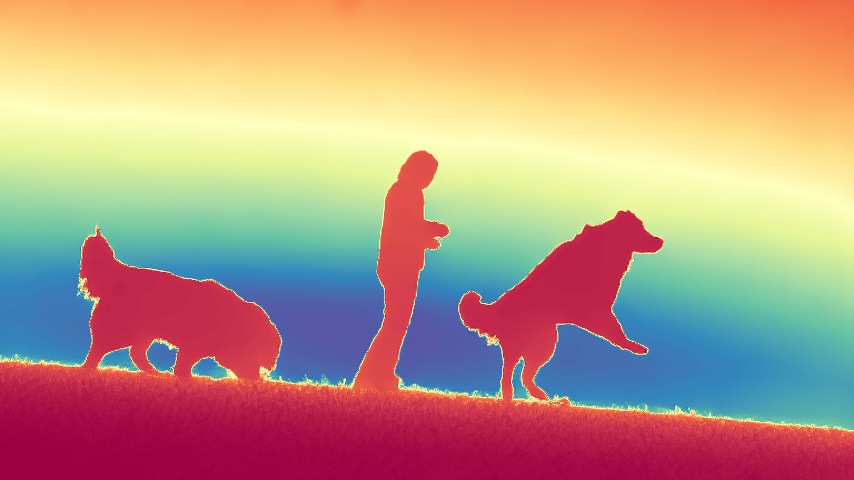} \\
\midrule
\raisebox{2.2\normalbaselineskip}[0pt][0pt]{\rotatebox[origin=c]{90}{Input}} &
\includegraphics[width=0.2\textwidth]{figures/depth/input_goat_00043.jpg} &
\includegraphics[width=0.2\textwidth]{figures/depth/input_goat_00044.jpg} &
\includegraphics[width=0.2\textwidth]{figures/depth/input_goat_00045.jpg} &
\includegraphics[width=0.2\textwidth]{figures/depth/input_goat_00046.jpg} &
\includegraphics[width=0.2\textwidth]{figures/depth/input_goat_00047.jpg} \\
\raisebox{2.2\normalbaselineskip}[0pt][0pt]{\rotatebox[origin=c]{90}{Marigold}} &
\includegraphics[width=0.2\textwidth]{figures/depth/perframe_goat_00043.jpg} &
\includegraphics[width=0.2\textwidth]{figures/depth/perframe_goat_00044.jpg} &
\includegraphics[width=0.2\textwidth]{figures/depth/perframe_goat_00045.jpg} &
\includegraphics[width=0.2\textwidth]{figures/depth/perframe_goat_00046.jpg} &
\includegraphics[width=0.2\textwidth]{figures/depth/perframe_goat_00047.jpg} \\
\raisebox{2.2\normalbaselineskip}[0pt][0pt]{\rotatebox[origin=c]{90}{Ours}} &
\includegraphics[width=0.2\textwidth]{figures/depth/ours_goat_00043.jpg} &
\includegraphics[width=0.2\textwidth]{figures/depth/ours_goat_00044.jpg} &
\includegraphics[width=0.2\textwidth]{figures/depth/ours_goat_00045.jpg} &
\includegraphics[width=0.2\textwidth]{figures/depth/ours_goat_00046.jpg} &
\includegraphics[width=0.2\textwidth]{figures/depth/ours_goat_00047.jpg} \\
\midrule
\raisebox{2.2\normalbaselineskip}[0pt][0pt]{\rotatebox[origin=c]{90}{Input}} &
\includegraphics[width=0.2\textwidth]{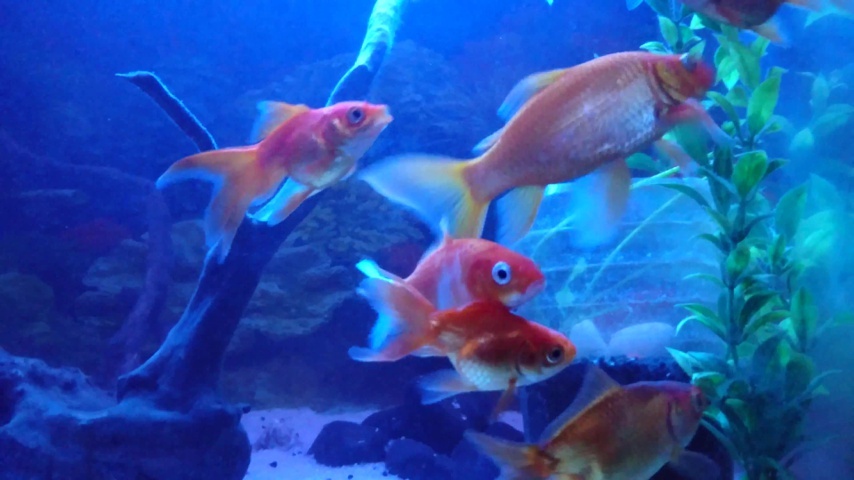} &
\includegraphics[width=0.2\textwidth]{figures/depth/input_gold-fish_00038.jpg} &
\includegraphics[width=0.2\textwidth]{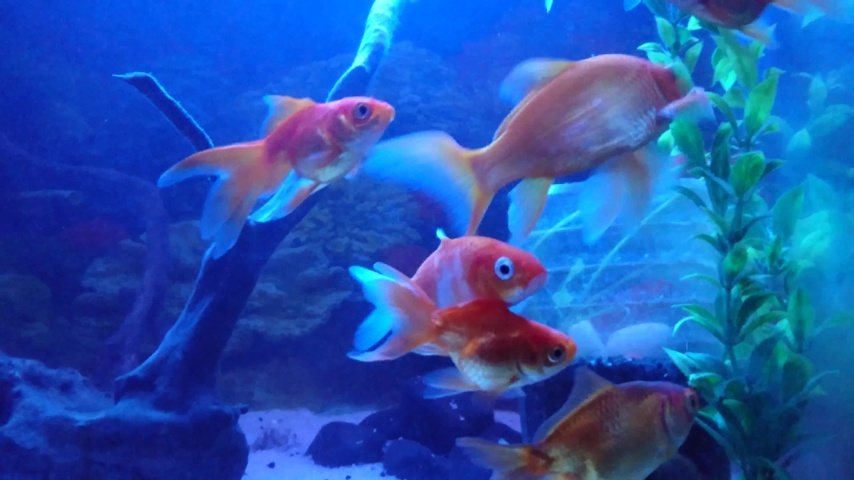} &
\includegraphics[width=0.2\textwidth]{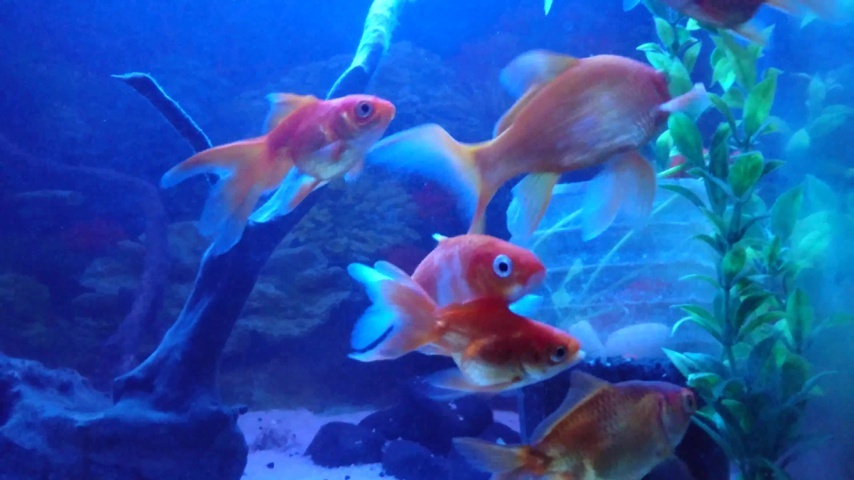} &
\includegraphics[width=0.2\textwidth]{figures/depth/input_gold-fish_00041.jpg} \\
\raisebox{2.2\normalbaselineskip}[0pt][0pt]{\rotatebox[origin=c]{90}{Marigold}} &
\includegraphics[width=0.2\textwidth]{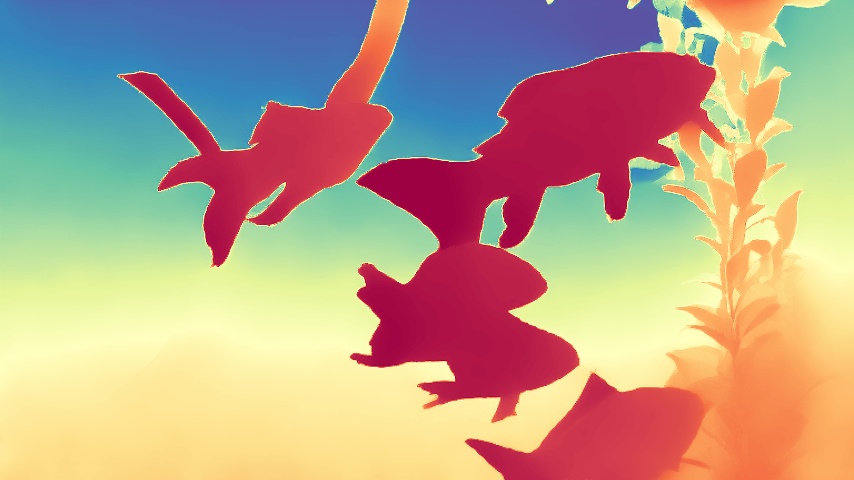} &
\includegraphics[width=0.2\textwidth]{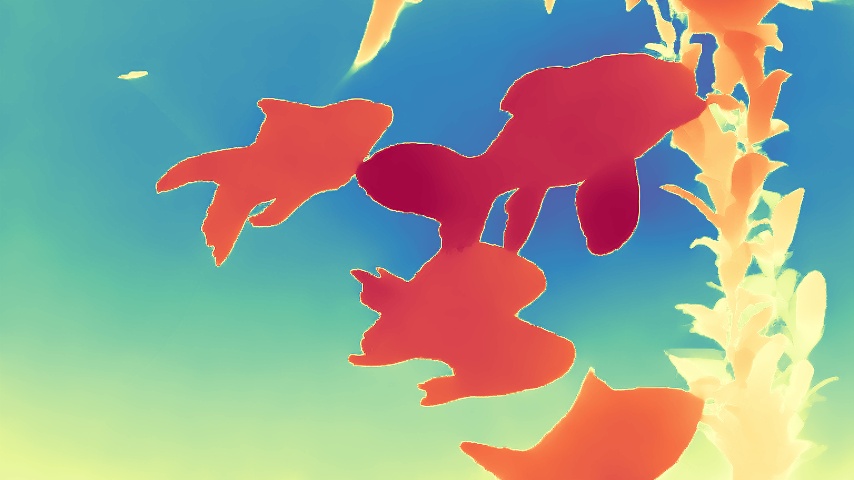} &
\includegraphics[width=0.2\textwidth]{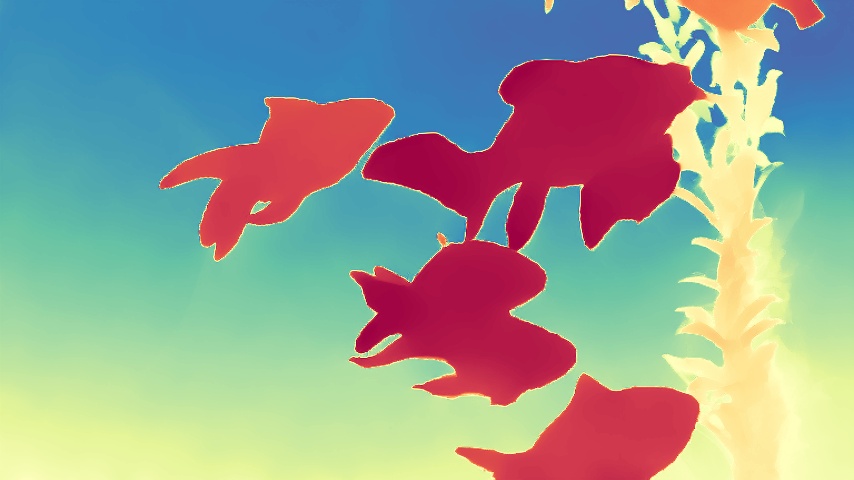} &
\includegraphics[width=0.2\textwidth]{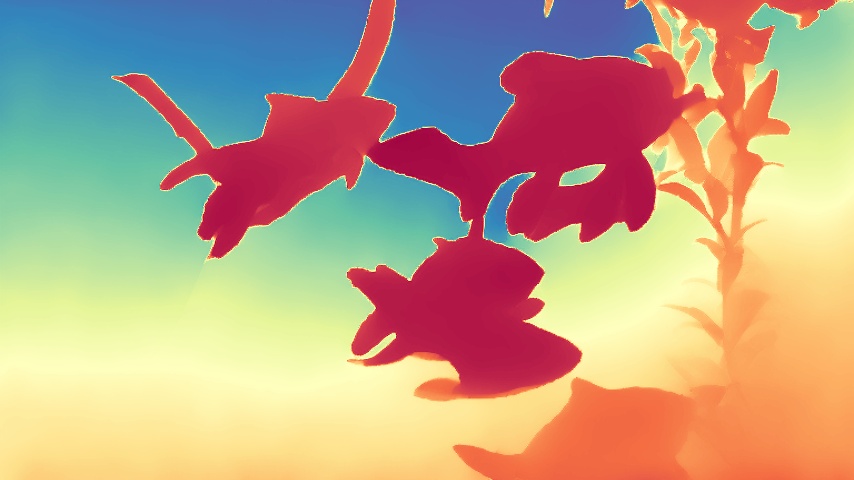} &
\includegraphics[width=0.2\textwidth]{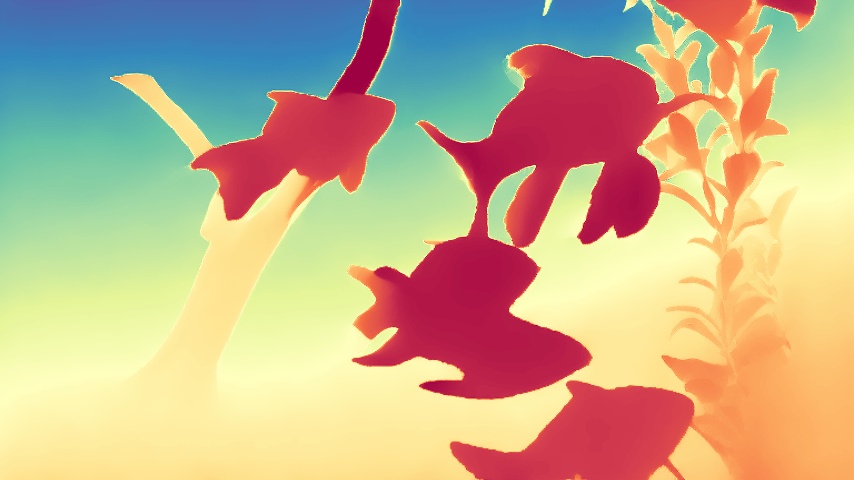} \\
\raisebox{2.2\normalbaselineskip}[0pt][0pt]{\rotatebox[origin=c]{90}{Ours}} &
\includegraphics[width=0.2\textwidth]{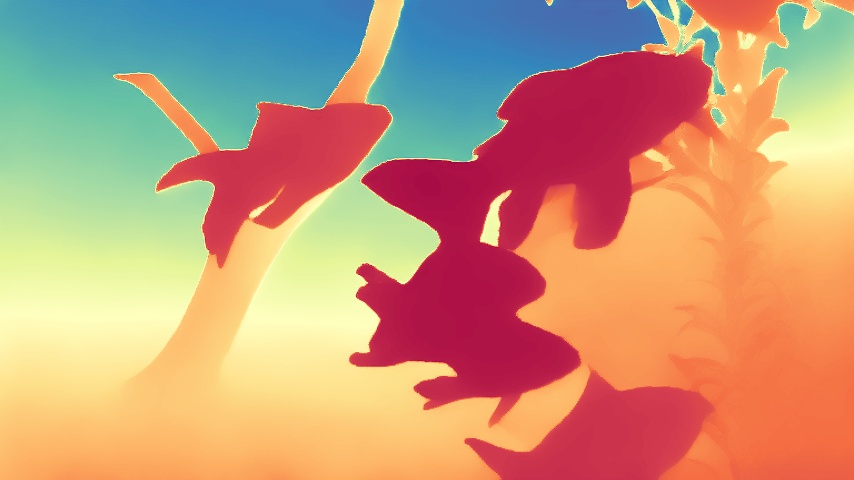} &
\includegraphics[width=0.2\textwidth]{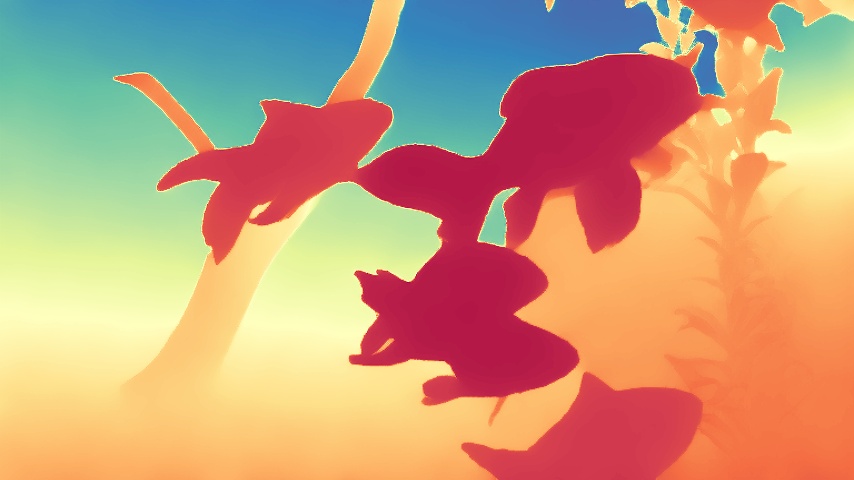} &
\includegraphics[width=0.2\textwidth]{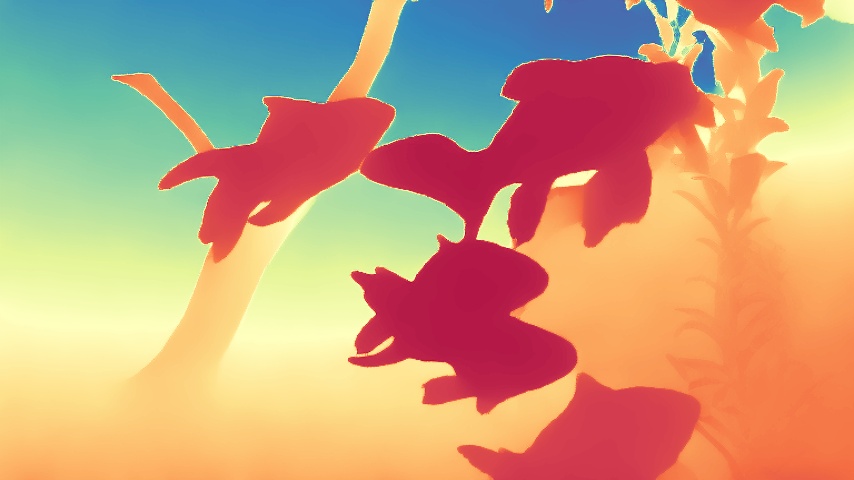} &
\includegraphics[width=0.2\textwidth]{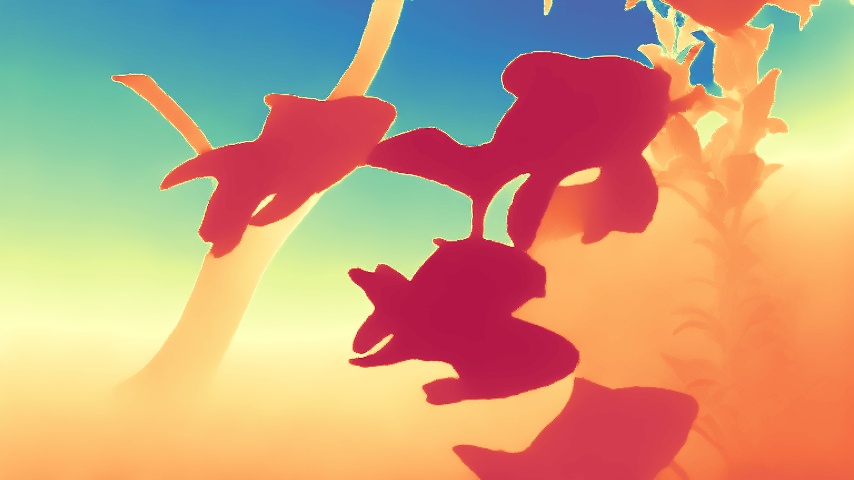} &
\includegraphics[width=0.2\textwidth]{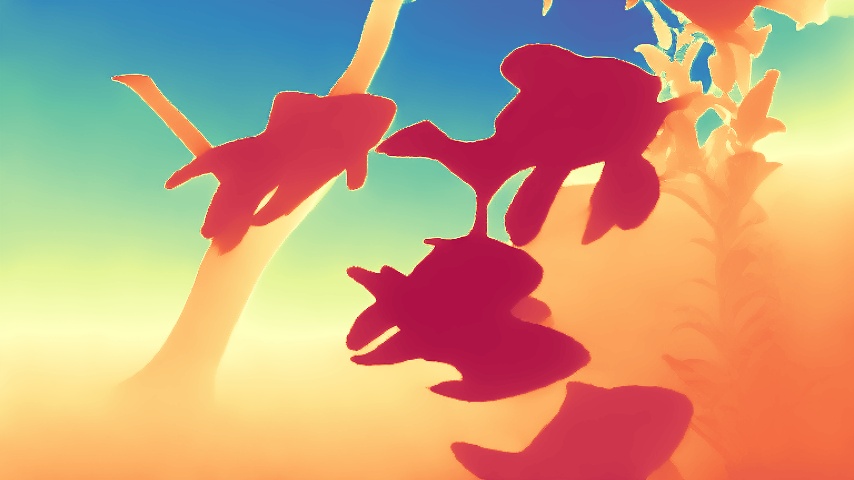} \\
\midrule
\raisebox{2.2\normalbaselineskip}[0pt][0pt]{\rotatebox[origin=c]{90}{Input}} &
\includegraphics[width=0.2\textwidth]{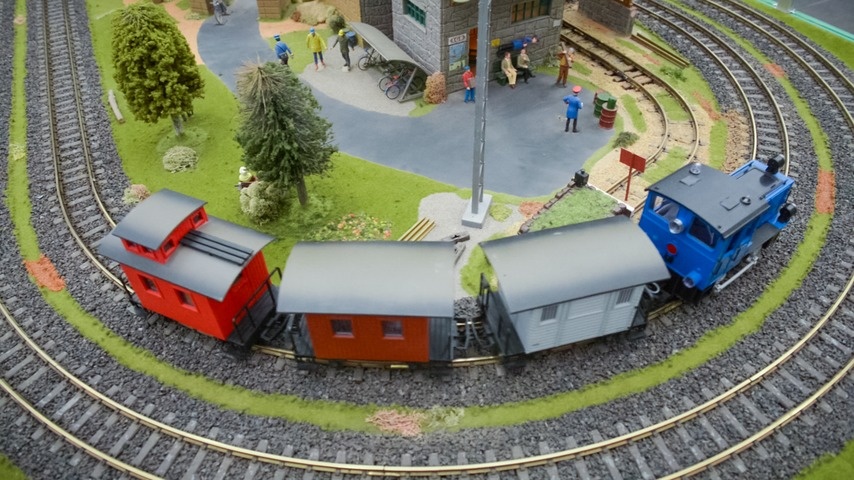} &
\includegraphics[width=0.2\textwidth]{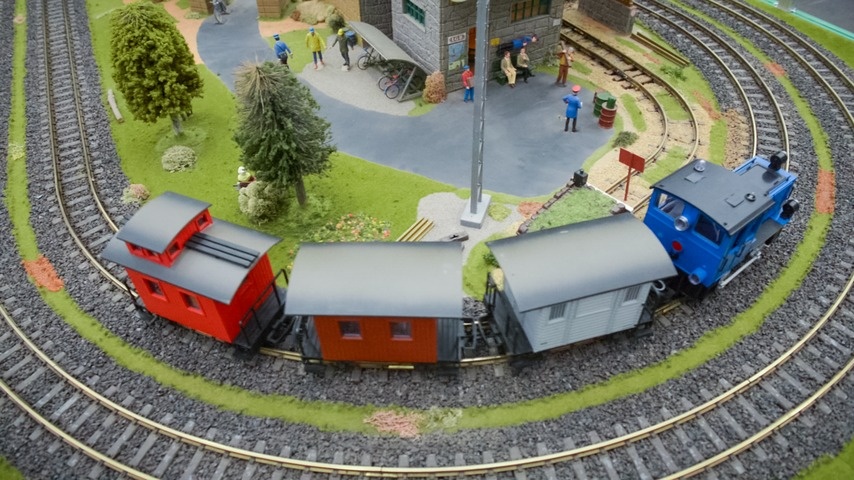} &
\includegraphics[width=0.2\textwidth]{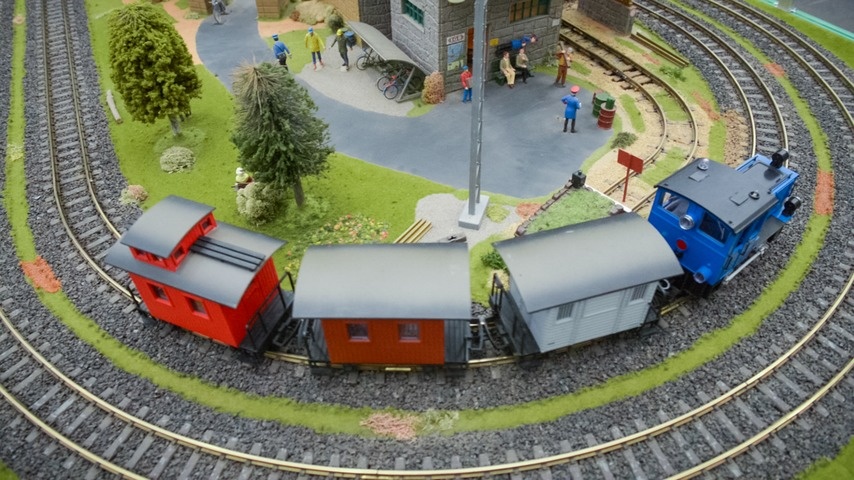} &
\includegraphics[width=0.2\textwidth]{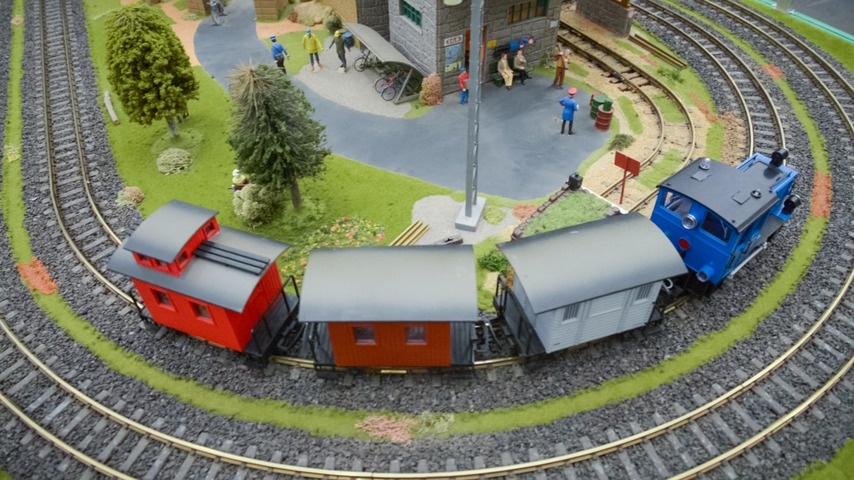} &
\includegraphics[width=0.2\textwidth]{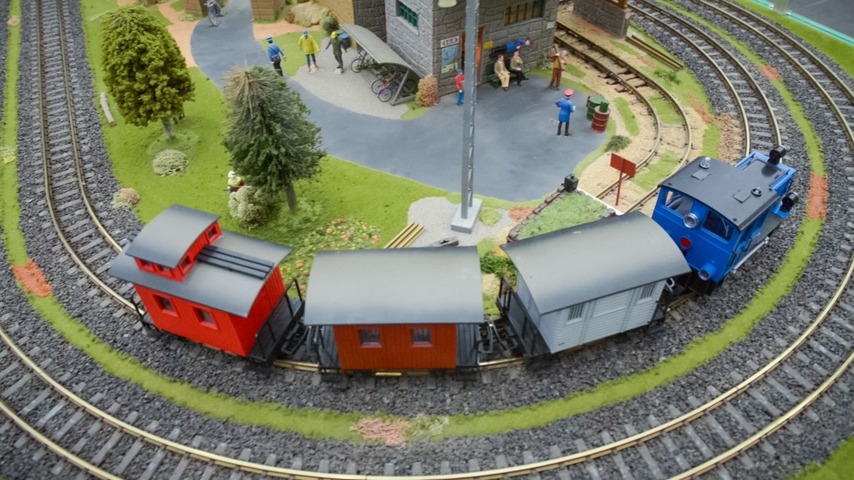} \\
\raisebox{2.2\normalbaselineskip}[0pt][0pt]{\rotatebox[origin=c]{90}{Marigold}} &
\includegraphics[width=0.2\textwidth]{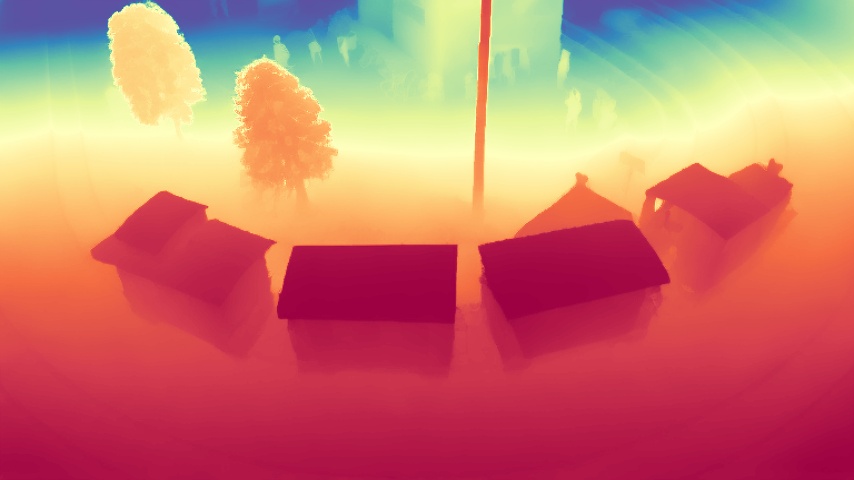} &
\includegraphics[width=0.2\textwidth]{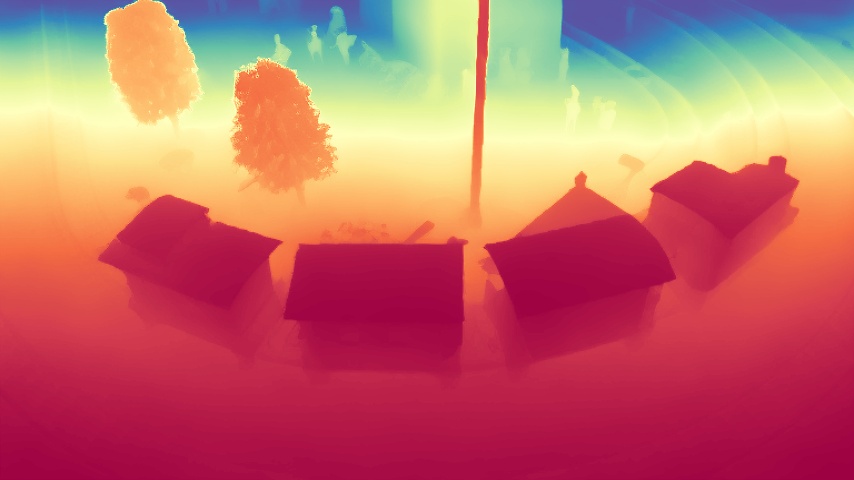} &
\includegraphics[width=0.2\textwidth]{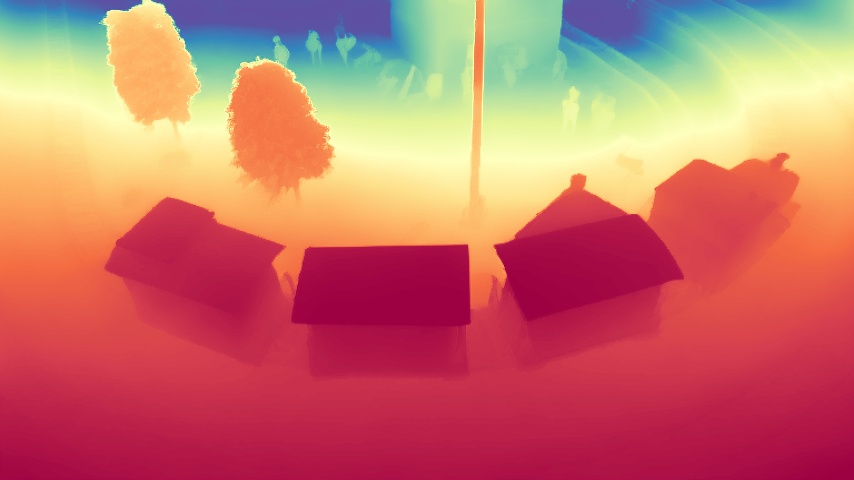} &
\includegraphics[width=0.2\textwidth]{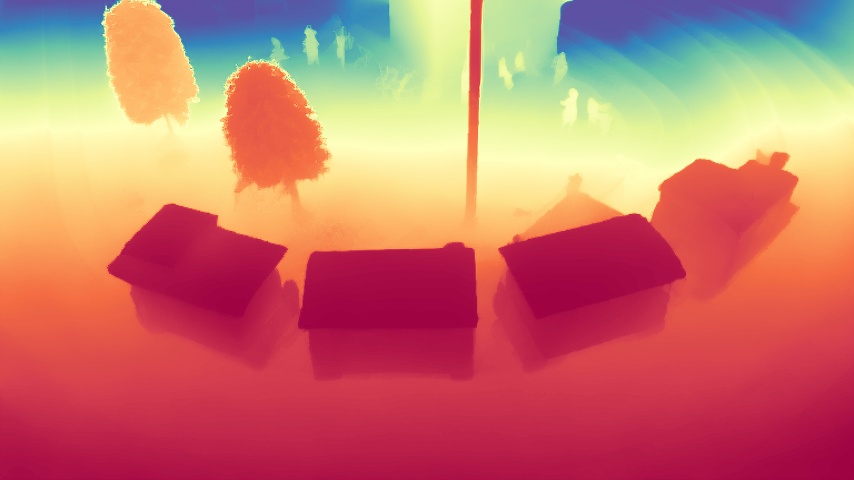} &
\includegraphics[width=0.2\textwidth]{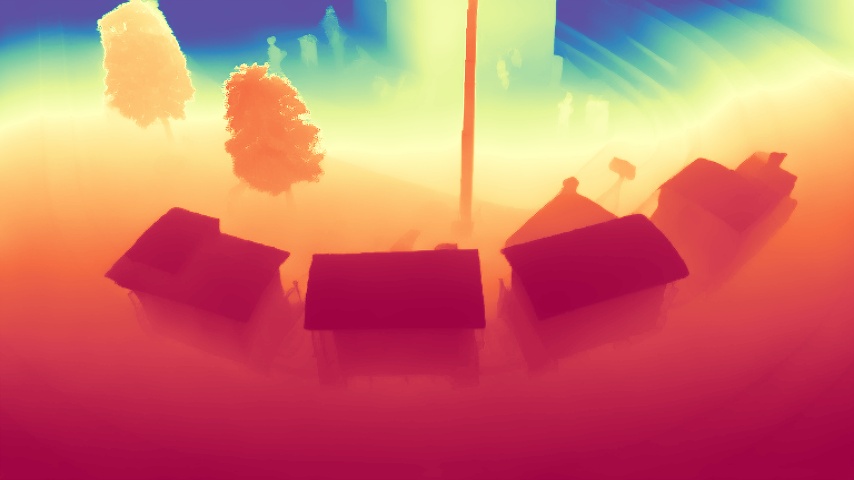} \\
\raisebox{2.2\normalbaselineskip}[0pt][0pt]{\rotatebox[origin=c]{90}{Ours}} &
\includegraphics[width=0.2\textwidth]{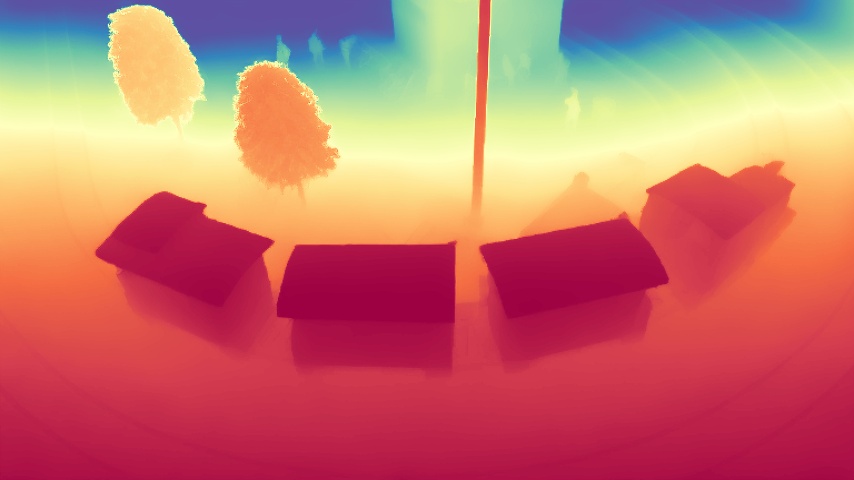} &
\includegraphics[width=0.2\textwidth]{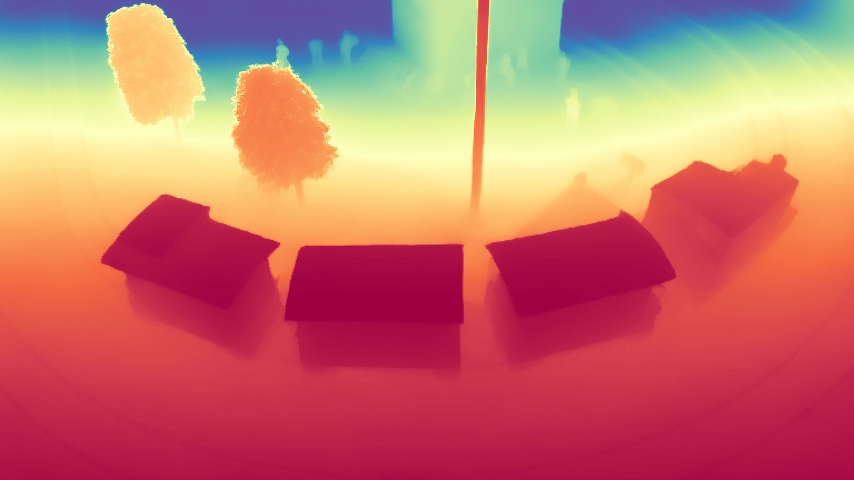} &
\includegraphics[width=0.2\textwidth]{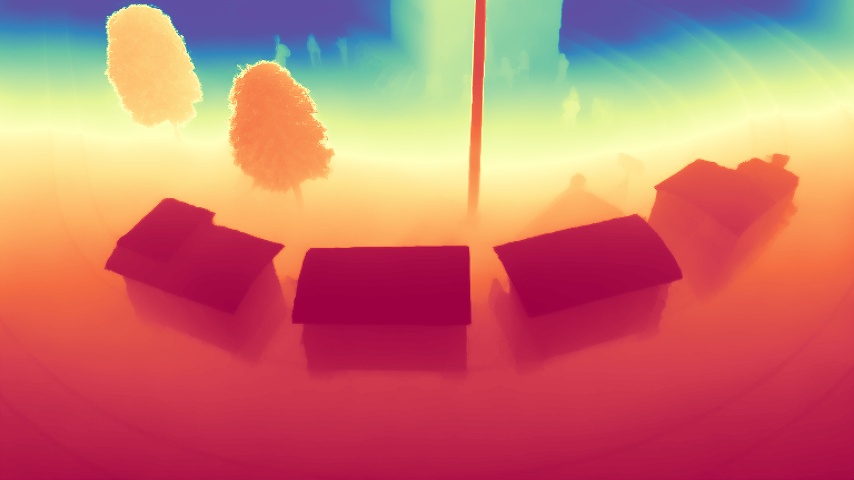} &
\includegraphics[width=0.2\textwidth]{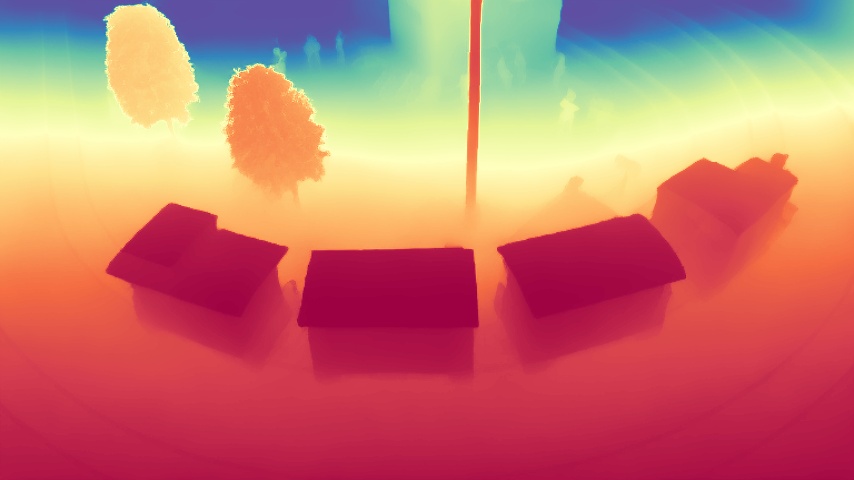} &
\includegraphics[width=0.2\textwidth]{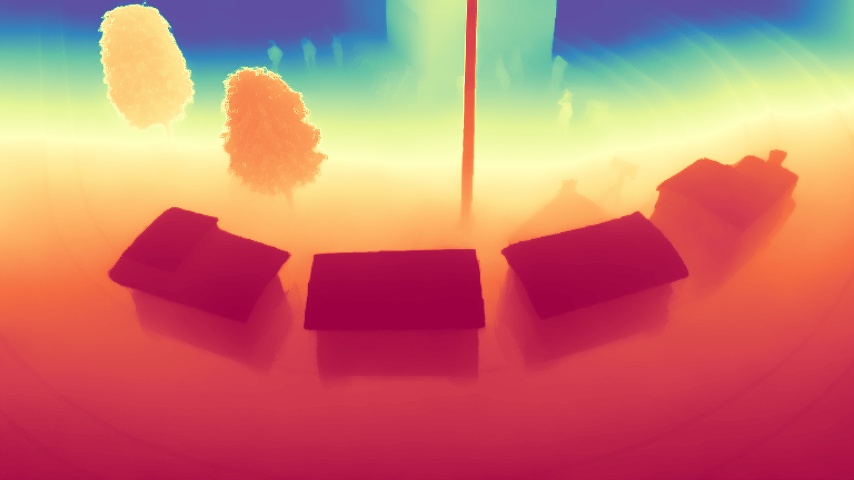} \\
\bottomrule
\end{tabular}%
}
\caption{\textbf{Applying our techniques to consistent video depth.} Integrating our proposed framework into Marigold~\citep{ke2024repurposing} helps improve the temporal consistency of video depth estimation.
}
\label{fig:depth_visual_suppl}
\end{figure*}

\subsection{Computational Complexity}
While our method focuses on zero-shot video restoration without additional training, it's important to consider the computational requirements in comparison to other approaches. \cref{tab:computational_complexity} provides an overview of the training time and GPU specifications for different methods, including ours.

As shown in the table, our method stands out by not requiring any training or fine-tuning, which significantly reduces the computational resources needed. This is in stark contrast to other methods that require multiple high-end GPUs and several days of training time.
For inference, our method introduces some computational overhead due to the hierarchical latent warping and hybrid token merging processes. However, this overhead is relatively small compared to the resources required for training or fine-tuning video models. Specifically, our method adds only approximately 6 seconds to the inference time of the base image diffusion model per frame.

\begin{table}[t]
    \centering
    \small
    \caption{\textbf{Training time and used devices for different methods.}}
    \label{tab:computational_complexity}
    \resizebox{1.0\columnwidth}{!} 
    {
    \begin{tabular}{l|cc}
    \toprule
    Method & Training time & GPU specs \\
    \midrule
    Shift-Net~\citep{yan2018shift} & Not reported & 8 NVIDIA A100-32G GPUs \\
    FMA-Net~\citep{youk2024fmanet} & Not reported & Not reported \\
    Upscale-A-Video~\citep{zhou2023upscale} & Not reported & 32 NVIDIA A100-80G GPUs  \\
    Ours & No training needed & - \\
    \bottomrule
    \end{tabular}%
    }
\end{table}

\subsection{Inference Time Comparison}
We report the inference time for processing 10 video frames at 854$\times$480 resolution on a single 4090 GPU in~\cref{tab:inference_time}. Our method adds a reasonable overhead compared to per-frame inference (around 26s for SDx4 and 63s for DiffBIR) while maintaining strong temporal consistency. This is notably more efficient than training-based methods like Upscale-A-Video, which requires 32 A100 GPUs and encounters out-of-memory (OOM) issues even during inference on our test setup. Furthermore, when applied to lightweight models like Marigold with 4-step sampling, our method achieves very fast inference at just 10 seconds total.

\begin{table}[t]
    \centering  
    \small
    \caption{\textbf{Inference time different methods.}}
    \label{tab:inference_time}
    \begin{tabular}{lc}
    \toprule
    Method & Inference time \\
    \midrule
    VidToMe~\cite{li2024vidtome} & 1m 49s \\
    FMA-Net~\cite{youk2024fmanet}	& 4.7s \\
    SDx4~\cite{sdx4} per-frame & 41s \\
    SDx4~\cite{sdx4} + Ours	& 1m 7s \\
    DiffBIR~\cite{lin2024diffbir} per-frame& 1m 17s \\
    DiffBIR~\cite{lin2024diffbir} + Ours & 2m 20s \\
    Shift-Net~\cite{yan2018shift} & 12.7s \\
    Marigold~\cite{ke2024repurposing} (4-step) + Ours & 10s \\
    Upscale-a-Video~\cite{zhou2023upscale} & OOM \\
    \bottomrule
    \end{tabular}%
\end{table}

\subsection{Additional Ablation Studies}
\paragraph{Comparison of temporal profiles.}
The comparisons in \cref{fig:ablation_temporal} also indicate that our results are smoother, demonstrating better temporal stability.
\begin{figure}[t]
\centering
\resizebox{1.0\columnwidth}{!} 
{
\includegraphics[width=\textwidth]{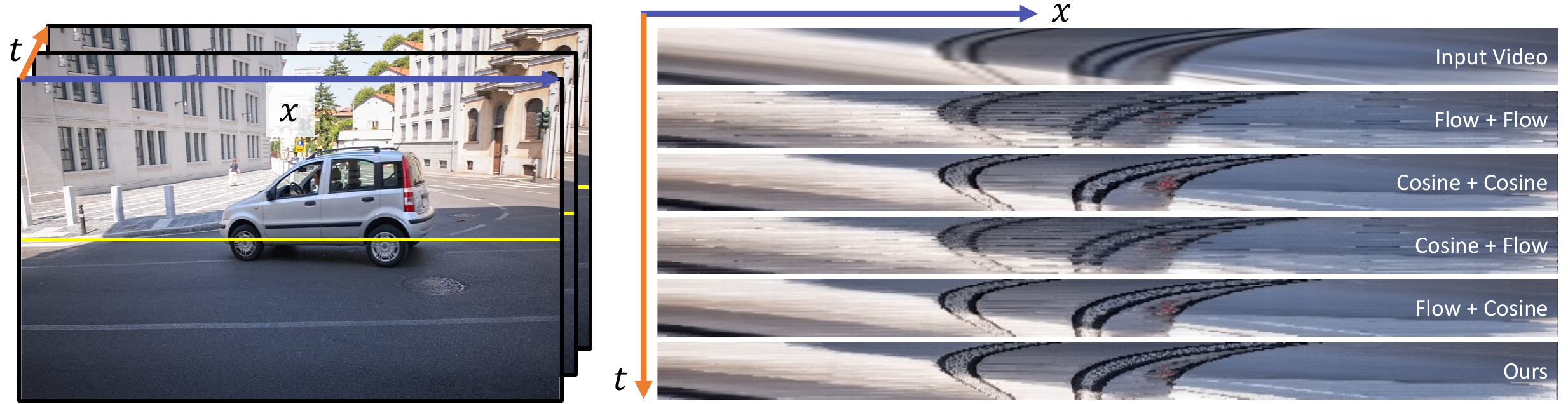}
}
\caption{\textbf{Comparison of temporal profile.} We examine a row of pixels and track changes over time. The profiles from {Flow + Flow} and {Cosine + Flow} methods exhibit noise, indicating flickering artifacts. The {Cosine + Cosine} method shows smoother profiles but contains some discontinuities. {Flow + Cosine} demonstrates improved consistency but retains some distortions. Utilizing flow, cosine, and spatial-aware techniques, our method achieves the most seamless and consistent transitions, effectively minimizing artifacts.
}
\label{fig:ablation_temporal}
\end{figure}

\paragraph{Token Unmerging Strategies.}
We experimented with two unmerging strategies: averaging paired tokens and direct replacement with keyframe tokens. \cref{tab:token_unmerging} shows the results of these experiments on the Vid4 x4 SR task.
As shown in the table, the replacement method outperforms averaging in terms of LPIPS, indicating better perceptual quality. Our experiments consistently showed that averaging tends to produce blurrier outputs in restoration tasks. Based on these results, we adopted the replacement-based unmerging process in our final model, as it preserves more details and leads to sharper outputs.

\begin{table}[t]
    \centering
    \small
    \caption{\textbf{Quantitative comparisons of different unmerging methods on Vid4 x4 SR task.}}
    \label{tab:token_unmerging}
    \begin{tabular}{l|c}
    \toprule
    Unmerging Method & LPIPS $\downarrow$ \\
    \midrule
    Averaging & 0.337 \\
    Replacement & 0.329 \\
    \bottomrule
    \end{tabular}%
\end{table}

\paragraph{Limitations: Extreme Degradation}
Extreme degradation (\emph{e.g.}, 32$\times$ super-resolution) or overly detailed facial features may yield unsatisfactory results (\cref{fig:failure}).
However, our framework's adaptability allows the incorporation of future, more powerful image-based diffusion models. Future improvements will focus on refining keyframe selection, stabilizing decoder output across LDM architectures, and enhancing extreme degradation handling. These aim to improve practical application and mitigate flickering issues inherent in LDM decoders.

\begin{figure}[t]
\centering
\footnotesize
\setlength{\tabcolsep}{1pt}
\renewcommand{\arraystretch}{1}
\resizebox{\columnwidth}{!}{%
\begin{tabular}{ccc}
\includegraphics[width=0.3\textwidth]{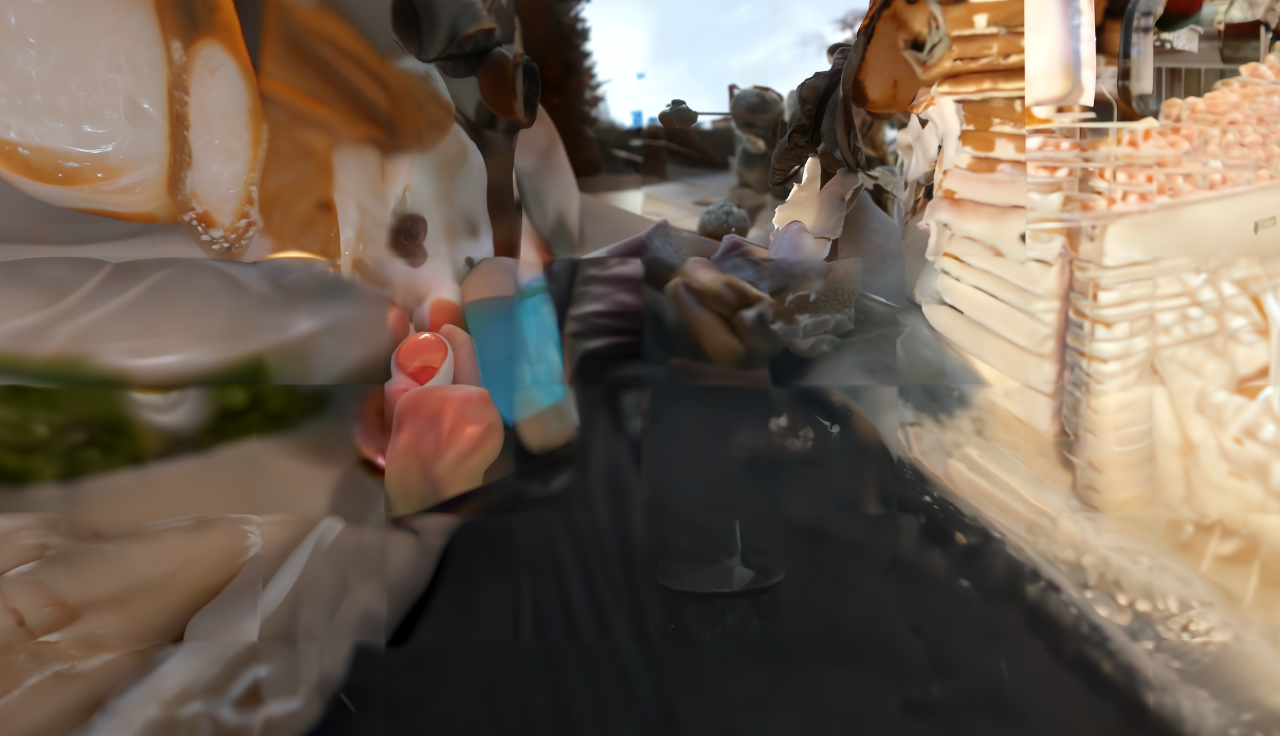} &
\includegraphics[width=0.3\textwidth]{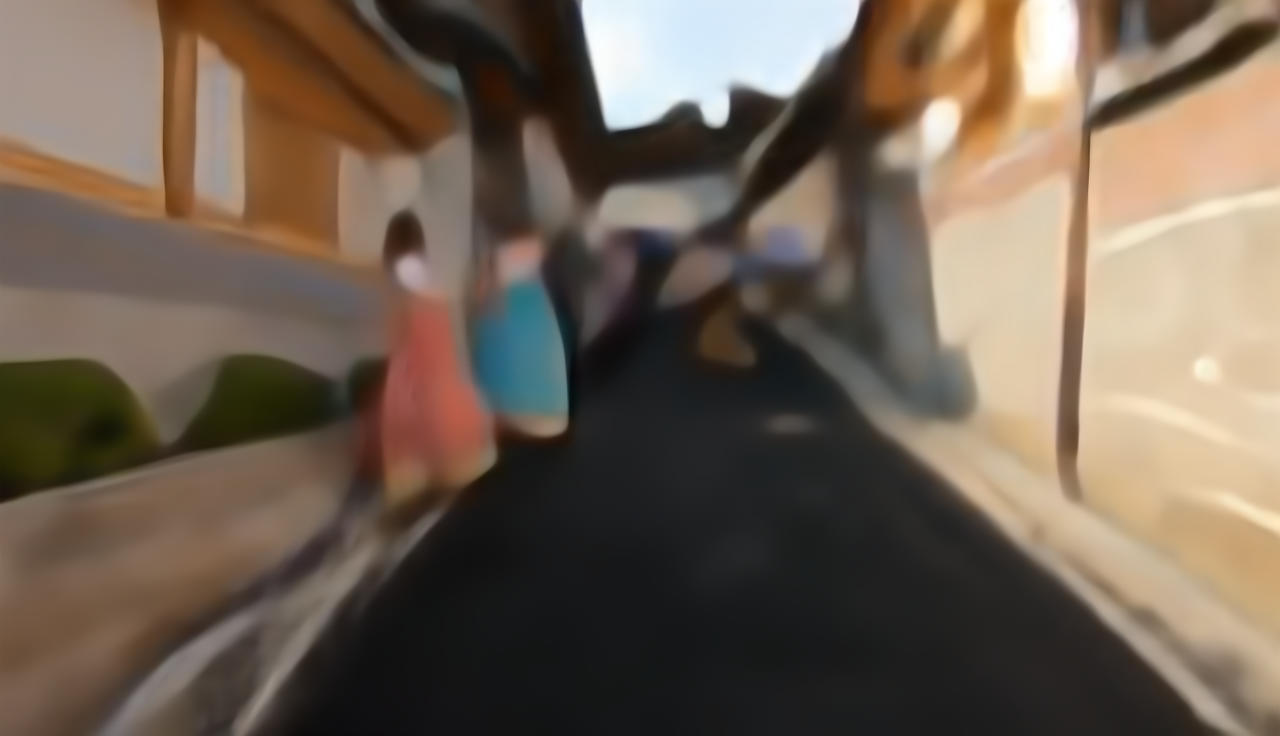} &
\includegraphics[width=0.3\textwidth]{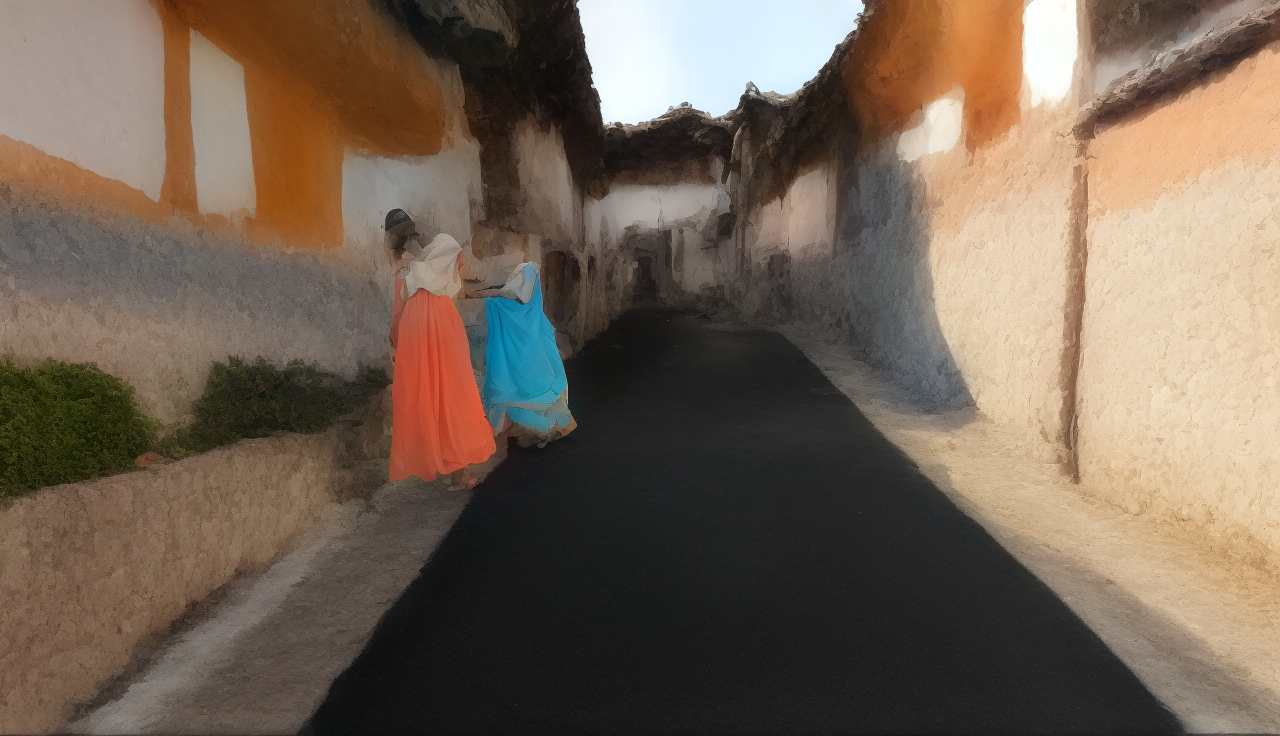} \\
\text{(a) DDNM} & \text{(b) FMA-Net} & \text{(c) Ours} \\
\end{tabular}%
}
\caption{\textbf{Failure case under 32x SR.} Most methods fail under this extreme degradation. However, if more powerful image-based diffusion models emerge in the future, our method can be easily adapted, offering greater potential to achieve this task.}
\label{fig:failure}
\end{figure}
 \fi

\end{document}